 \documentclass[final]{IEEEtran}

\usepackage{subfigure}
\usepackage{algorithm}  
\usepackage{algorithmic} 

\usepackage{xr,color}
\usepackage{bm}
\usepackage{multirow}

\usepackage{amsmath,amsfonts,amssymb,amsthm,graphicx,graphics,subfigure,epsfig,enumerate,bm}
 
\usepackage{graphicx}

\usepackage{lipsum}

\definecolor{yc}{RGB}{255,0,0}
\definecolor{hf}{RGB}{125,0,0}
\definecolor{yl}{RGB}{0,0,200}

\let\hat\widehat
\let\tilde\widetilde

\newcommand{\ba}{\bm{a}}

\newcommand{\be}{\bm{e}}

\newcommand{\bs}{\bm{s}}

\newcommand{\bu}{\bm{u}}
\newcommand{\bv}{\bm{v}}
\newcommand{\bw}{\bm{w}}
\newcommand{\bx}{\bm{x}}

\newcommand{\bz}{\bm{z}}

\newcommand{\bA}{\bm{A}}
\newcommand{\bB}{\bm{B}}
\newcommand{\bC}{\bm{C}}

\newcommand{\bI}{\bm{I}}

\newcommand{\bM}{\bm{M}}

\newcommand{\bP}{\bm{P}}

\newcommand{\bR}{\bm{R}}

\newcommand{\bT}{\bm{T}}

\newcommand{\bV}{\bm{V}}
\newcommand{\bW}{\bm{W}}

\newcommand{\balpha}{\bm{\alpha}}

\theoremstyle{plain} \newtheorem{lemma}{\textbf{Lemma}} \newtheorem{theorem}{\textbf{Theorem}}
\newtheorem{corollary}{\textbf{Corollary}} \newtheorem{assumption}{\textbf{Assumption}}
 \newtheorem{definition}{\textbf{Definition}}
 \theoremstyle{definition}

\theoremstyle{remark}

\newtheorem*{theorem*}{\textbf{Theorem}}
\newtheorem*{lemma*}{\textbf{Lemma}}	

\usepackage{url}            
\usepackage{booktabs}       
\usepackage{amsfonts}       
\usepackage{nicefrac}       
\usepackage{microtype}      
\usepackage{cleveref}

\title{Guaranteed Recovery of One-Hidden-Layer Neural Networks via Cross Entropy}

\author{Haoyu Fu, Yuejie Chi, Yingbin Liang
	\thanks{H. Fu and Y. Liang are with Dept. of ECE, The Ohio State University, Columbus, OH 43210, USA. Emails: \{fu.436, liang.889\}osu.edu.}
	\thanks{Y. Chi is with Dept. of ECE, Carnegie Mellon University, Pittsburgh, PA 15213, USA. Email: yuejiechi@cmu.edu.}
	\thanks{The work of H. Fu and Y. Liang is supported in part by U.S. National Science Foundation under the grants CCF-1761506, CCF-1801855 and CCF-1900145. 
	The work of Y. Chi is supported in part by AFOSR under the grant FA9550-15-1-0205, by ONR under the grant N00014-18-1-2142, by ARO under the grant W911NF-18-1-0303, and by NSF under the grants CAREER ECCS-1818571, ECCS-1833553, CCF-1806154 and CCF-1901199.}
}

\begin{document}

\maketitle

\begin{abstract}
	We study model recovery for data classification, where the training labels are generated from a one-hidden-layer neural network with sigmoid activations, also known as a single-layer feedforward network, and the goal is to recover the weights of the neural network. We consider two network models, the fully-connected network (FCN) and the non-overlapping convolutional neural network (CNN). We prove that with Gaussian inputs, the empirical risk based on cross entropy exhibits strong convexity and smoothness {\em uniformly} in a local neighborhood of the ground truth, as soon as the sample complexity is sufficiently large. This implies that if initialized in this neighborhood, gradient descent converges linearly to a critical point that is provably close to the ground truth. Furthermore, we show such an initialization can be obtained via the tensor method. This establishes the global convergence guarantee for empirical risk minimization using cross entropy via gradient descent for learning one-hidden-layer neural networks, at the near-optimal sample and computational complexity with respect to the network input dimension without unrealistic assumptions such as requiring a fresh set of samples at each iteration.
\end{abstract}

\section{Introduction}

Neural networks have attracted a significant amount of research interest in recent years due to the success of deep neural networks \cite{lecun2015deep} in practical domains such as computer vision and artificial intelligence \cite{russakovsky2015imagenet,he2016deep,silver2016mastering}. However, the theoretical underpinnings behind such success remains mysterious to a large extent. Efforts have been taken to understand which classes of functions can be represented by deep neural networks \cite{cybenko1989approximation,hornik1989multilayer,barron1993universal,telgarsky2016benefits}, when  (stochastic) gradient descent is effective for optimizing a nonconvex loss function \cite{dauphin2014identifying}, and why these networks generalize well \cite{zhang2016understanding,bartlett2017spectrally,brutzkus2018sgd}.

One important line of research that has attracted extensive attention is the model-recovery problem, which is important for the network to generalize well \cite{mondelli2018connection}. Specifically, it is shown in \cite{mondelli2018connection} that in a model-recovery setting, a network cannot generalize well if the underlying parameters cannot be recovered accurately, therefore linking model recovery to generalization. In addition, the problem of model recovery provides a framework to leverage statistical nature of the input data in an intuitive manner, which allows shedding more light to the understanding of optimization of complex neural networks.

Let the training samples $ (\bm{x}_i,y_i ) \sim (\bm{x}, y) $, $i=1,\ldots,n$, be generated independently and identically distributed (i.i.d.) from a distribution $\mathcal{D}$ based on a neural network model with the ground truth parameter $\bW^{\star}$, and the goal is to recover $\bW^{\star}$ using the training samples given the network architecture. Consider a network whose output is given as $H(\bm{W}^{\star}, \bm{x})$. Previous studies along this topic can be mainly divided into two cases of data generations, with the input $\bx\in\mathbb{R}^d$ being drawn from the Gaussian distribution.
\begin{itemize}
\item {\em Regression}, where each sample $y\in\mathbb{R}$ is generated as 
$$ y =  H(\bm{W}^{\star}, \bm{x}).$$ 
This type of regression problem has been studied in various settings. In particular, \cite{soltanolkotabi2017learning} studied the single-neuron model under the Rectified Linear Unit (ReLU) activation, \cite{zhong17a} studied the one-hidden-layer multi-neuron network model, and \cite{li2017convergence} studied a two-layer feedforward network with ReLU activations and identity mapping. 
\item {\em Classification}, where a label $y\in\{0,1\}$ is drawn according to the conditional distribution 
$$ \mathbb{P}(y=1 |\bm{x}) = H(\bm{W}^{\star}, \bm{x}).$$ 
Such a problem has been studied in \cite{mei2016landscape} when the network contains only a single neuron. 
\end{itemize}
For both cases, previous studies attempted to recover $\bm{W}^{\star}$, by minimizing an empirical loss function using the squared loss, i.e. $\min_{\bm{W}} \frac{1}{n}\sum_{i=1}^n (y_i -H(\bm{W}, \bm{x}_i) )^2$, given the training data.  Two types of statistical guarantees were provided for such model recovery problems using the squared loss. More specifically, \cite{zhong17a} showed that in the local neighborhood of the ground truth $\bm{W}^{\star}$, the {\em empirical} loss function is strongly convex for each {\em given} point under {\em independent} high probability event, which implies that {\em fresh samples} are required at {\em every} iteration for gradient descent to  converge linearly with well-designed initializations. On the other hand, studies such as \cite{mei2016landscape} established strong convexity in the entire local neighborhood of the ground truth in a uniform sense, so that resampling per iteration is not needed for gradient descent to have guaranteed linear convergence as long as it enters such a local neighborhood. Here, one weakness of the pointwise strong convexity in \cite{zhong17a}, compared to the uniform strong convexity in \cite{mei2016landscape}, is that independent fresh samples are required at each iteration to guarantee the linear convergence of gradient descent. Consequently, the sample complexity of \cite{zhong17a} grows with respect to the recovery accuracy $\epsilon$, typically with an extra factor of $\log (1/\epsilon)$ under linear convergence, which can be large when the desired accuracy is high. Therefore, the latter type of uniform strong convexity {\em without requiring per-iteration resampling} is much stronger and more desirable. 


\begin{figure*}[t]
\begin{center}
		\begin{tabular}{cc}
			\includegraphics[width=0.4\textwidth]{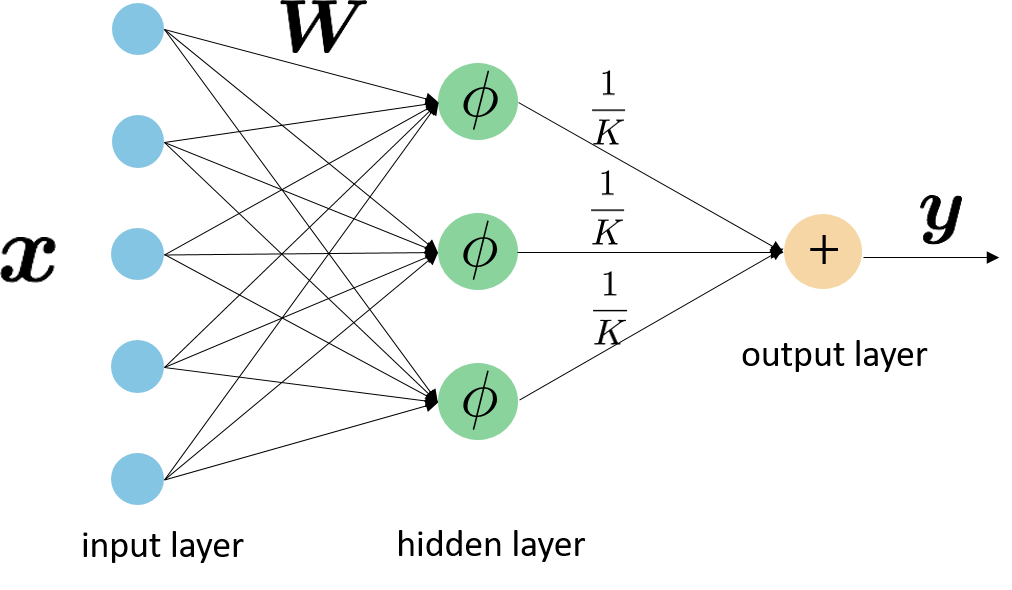} & \quad \includegraphics[width=0.3\textwidth]{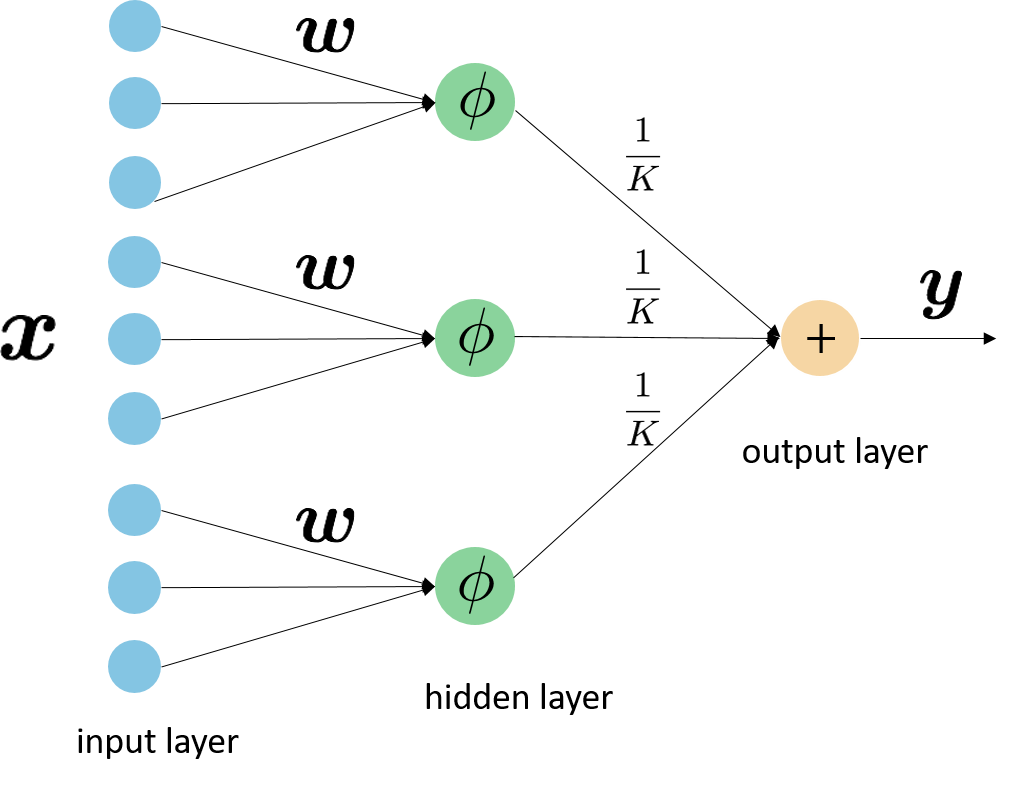} \\
			\hspace{-0.2in}(a) FCN & \quad (b) CNN
		\end{tabular}
\end{center}
	\caption{Illustration of two types of one-hidden-layer neural networks considered in this paper: (a) a fully-connected network (FCN); (b) a non-overlapping convolutional neural network (CNN). }\label{Fig:1NN_illustration}
\end{figure*}

In this paper,  we focus on the classification setting by minimizing the empirical loss using the cross entropy objective, which is a popular choice in training practical neural networks. The geometry as well as the optimization of the model recovery problem based on the cross-entropy loss function have not yet been understood even for one-hidden-layer networks. The main focus of this paper is to develop technical analysis for guaranteed model recovery under the challenging cross-entropy loss function for the classification problem for two types of one-hidden-layer network structures.



\subsection{Problem Formulation} \label{sec:formulation}
 
We consider two popular types of one-hidden-layer nonlinear neural networks illustrated in Fig.~\ref{Fig:1NN_illustration}, i.e., a Fully-Connected Network (FCN) \cite{zhong17a} and a non-overlapping Convolutional Neural Network (CNN) \cite{brutzkus2017globally}. For both cases, we let $\bm{x}\in\mathbb{R}^d$ be the input, $K\geq 1$ be the number of neurons, and the activation function be the sigmoid function
$$ \phi\left(x \right) = \frac{1}{1+\exp\left( -x\right) } .$$
\begin{itemize}
	\item {\em FCN:} the network parameter is $\bW= [\bw_{1},\cdots,\bw_{K}]\in \mathbb{R}^{d\times K}$, and
	\begin{equation}\label{eq:FNN:Classification_model}
	 H_{\mathrm{FCN}}\left(\bW ,\bx \right) = \frac{1}{K} \sum_{k=1}^K \phi(\bw_k^{ \top}\bx).
	\end{equation}
		 
	\item{\em Non-overlapping CNN:} for simplicity we let $d= m  K$ for some integers $m$. Let $\bw \in\mathbb{R}^m$ be the network parameter, and the $k$th stride of $\bm{x}$ be given as $\bx^{\left(k \right) } = \left[x_{m\left(k-1 \right)+1 },\cdots x_{m\cdot k} \right]^{\top}\in \mathbb{R}^{m}$.  Then, 
	\begin{equation}\label{eq:CNN:Classification_model}
	H_{\mathrm{CNN}}\left(\bw,\bx \right) = \frac{1}{K} \sum_{k=1}^K \phi(\bw^{\top}\bx^{\left(k \right) }).
	\end{equation}
\end{itemize}
The non-overlapping CNN model can be viewed as a highly structured instance of the FCN, where the weight matrix can be written as:
\begin{equation*}
\bm{W}_{\mathrm{CNN}} = \begin{bmatrix}
\bw &  \bm{0}   & \ldots & \bm{0} \\
\bm{0} & \bw & \ldots & \bm{0} \\
\vdots & \vdots & \ddots & \vdots \\
\bm{0} & \bm{0} & \ldots & \bm{w} \\
\end{bmatrix} \in \mathbb{R}^{d\times K}.
\end{equation*}


In a model recovery setting, we are given $n$ training samples $\left\lbrace  \left(\bx_{i},y_{i} \right) \right\rbrace_{i=1}^{n} \sim (\bx, y)$ that are drawn i.i.d. from certain distribution regarding the ground truth network parameter $\bW^{\star}$ (or resp. $\bw^{\star}$ for CNN). Suppose the network input $\bx\in\mathbb{R}^d$ is drawn from a standard Gaussian distribution $\bx \sim \mathcal{N}(\bm{0},\bI_d)$. This assumption has been used a lot in previous literature \cite{soltanolkotabi2017learning,oymak2018learning,brutzkus2017globally,du2017convolutional}, to name a few. Then, conditioned on $\bx\in\mathbb{R}^d$, the output $y$ is mapped to $\{0,1\}$ via the output of the neural network, i.e., 
\begin{align} \label{eq:label_generation}
	\mathbb{P}\left(y=1|\bx \right) = H\left(\bW^{\star},\bx \right).
\end{align}

 Our goal is to recover the network parameter, i.e., $\bW^{\star}$. One natural choice is to maximize the log-likelihood function, which turns out to be equivalent to minimizing
 \begin{equation}\label{Eq:Object}
 	f_n (\bW) = \frac{1}{n}\sum_{i=1}^{n} \ell\left(\bW; \bx_{i},y_{i} \right),
 \end{equation}
 where $\ell\left(\bW;\bx, y \right)$ is the cross-entropy loss function, i.e.,
 \begin{align}
 &\ell\left(\bW; \bx,y \right) \notag \\
 &= -   y \cdot \log\left(H\left(\bW,\bx \right)  \right) - ( 1-y) \cdot \log\left(1- H\left(\bW,\bx \right)\right)	,
 \end{align}
 where $H(\bW,\bx)$ can subsume either $H_{\mathrm{FCN}}$ or $H_{\mathrm{CNN}}$. Although the squared loss has been used in \cite{mei2016landscape} to study the classification problem with a single neuron, the cross-entropy loss is a more natural and popular choice in practice for classification data, due to its natural connection to the principle of maximum likelihood estimation.

\subsection{Our Contributions}



Considering the multi-neuron classification problem with either FCN or CNN, the main contributions of this work are summarized as follows. Throughout the discussions below, we assume the number $K$ of neurons is a constant, and state the scaling only in terms of the input dimension $d$ and the number $n$ of samples.  
\begin{itemize}
	\item {\em Uniform local strong convexity:} If the input is Gaussian, the empirical risk function $f_n(\bm{W}) $ is {\em uniformly} strongly convex in a local neighborhood of the ground truth $\bm{W}^{\star}$ as soon as the sample size $n= O(d   \log^2 d)$. 
	\item {\em Statistical and computational rate of gradient descent:} consequently, if initialized in this neighborhood, gradient descent converges linearly to a critical point  (which we show to exist). Due to the nature of quantized labels here, the recovery of the ground truth is only up to certain statistical accuracy. In particular, gradient descent finds the critical point $\hat{\bW}_n$ with a computation cost of $O (nd \log ( 1/\epsilon )    )$, where $\epsilon$ denotes the numerical accuracy and $\hat{\bW}_n$ converges to $\bW^{\star}$ at a rate of $O (\sqrt{d \log  n/n} )$ in the Frobenius norm.

	
\item {\em Tensor initialization:} We adopt the tensor method proposed in \cite{zhong17a}, and show that it provably provides an initialization in the neighborhood of the ground truth both for FCN and CNN. In particular, we strengthened the guarantee of the tensor method by replacing the homogeneous assumption on activation functions in \cite{zhong17a} by a mild condition on the curvature of activation functions around $\bW^{\star}$, which holds for a larger class of activation functions including {\em sigmoid} and {\em tanh}.
\end{itemize}

The cross-entropy loss is much more challenging to analyze than the squared loss, e.g., its gradient and Hessian take much more complicated forms compared with the squared loss; moreover, it is hard to control the values of gradient and Hessian due to the saturation phenomenon, i.e., when $H\left(\bW,\bx \right) $ approaches $0$ or $1$.
In order to establish the uniform local strong convexity property for the cross-entropy loss, we first show the {\em population} loss is smooth regarding to $\bW^{\star}$. Such a property was also established in \cite{zhong17a} for the squared loss. However, considering the special form of Hessian under the cross-entropy loss, we need to apply Taylor's approximation together with
certain probabilistic upper bounds to control the value of Hessian, and obtain the smooth property. Network-specific quantities to capture the local geometry of the population loss at $\bW^{\star}$ for FCN and CNN are derived, which imply that the geometry of CNN is more benign than FCN, corroborated by the numerical experiments. 

Beyond these two steps, the additional uniform concentration property of the Hessian (Lemma~\ref{Uniform_Convergence}) is of key importance for us to obtain the uniform local strong convexity of the {\em empirical} loss. To show the uniform concentration of the Hessian, we successfully apply a type of covering argument. Different from the arguments in \cite{mei2016landscape}, which deal with the squared loss and are facilitated by certain nice assumptions on the activation functions, the cross-entropy loss is more difficult to apply the covering argument, e.g., both the gradient and Hessian no longer have a deterministic
upper bound. Hence, we exploit the property of the sigmoid activation to show that the gradient and the Hessian of the cross-entropy loss are upper bounded with high
probability in order to establish the uniform concentration property.

To the best of our knowledge, combining the analysis of gradient descent and initialization, this work provides the first globally convergent algorithm for the recovery of one-hidden-layer neural networks using the {\em cross-entropy} loss function.




\subsection{Related Work}

Due to the scope, we focus on the most relevant literature on theoretical and algorithmic aspects of learning shallow neural networks via nonconvex optimization. The parameter recovery viewpoint is relevant to the success of nonconvex learning in signal processing problems such as matrix completion, phase retrieval, blind deconvolution, dictionary learning and tensor decomposition \cite{sun2016guaranteed}\nocite{chen2018harnessing,candes2014wirtinger,ge2017optimization,ge2016matrix,sun2015complete,bhojanapalli2016global}--\cite{ma2017implicit}, to name a few; see also the overview article \cite{chi2018nonconvex}. The statistical model for data generation effectively removes worst-case instances and allows us to focus on average-case performance, which often possess much benign geometric properties that enable global convergence of simple local search algorithms.
 
 The studies of one-hidden-layer network model can be further categorized into two classes, landscape analysis and model recovery. In the landscape analysis, it is known that if the network size is large enough compared to the data input, then there are no spurious local minima in the optimization landscape, and all local minima are global \cite{soltanolkotabi2017theoretical,boob2017theoretical,safran2016quality,nguyen2017loss}. For the case with multiple neurons ($2\leq K \leq d$) in the under-parameterized setting, the work of Tian \cite{tian2017analytical} studied the landscape of the population squared loss surface with ReLU activations. In particular, there exist spurious bad local minima in the optimization landscape \cite{ge2017learning,safran2017spurious} even at the population level. Zhong et. al. \cite{zhong17a} provided several important geometric characterizations for the regression problem using a variety of activation functions and the squared loss. 
 
In the model recovery problem, the number of neurons is smaller than the input dimension, and all the existing works discussed below assumed the squared loss and (sub-)Gaussian inputs. In the case with a single neuron ($K=1$), \cite{soltanolkotabi2017learning} showed that gradient descent converges linearly when the activation function is ReLU, with a zero initialization, as long as the sample complexity is $O(d)$ for the regression problem. When the activation function is quadratic, \cite{chen2018gradient} shows that randomly initialized gradient descent converges fast to the global optimum at a near-optimal sample complexity. On the other hand, \cite{mei2016landscape} showed that when $\phi(\cdot)$ has bounded first, second and third derivatives, there is no other critical points than the unique global minimum (within a constrained region of interest), and (projected) gradient descent converges linearly with an arbitrary initialization, as long as the sample complexity is $O(d\log^{2} d)$ for the classification problem. Moreover, in the case with multiple neurons, \cite{oymak2018learning} showed that projected gradient descent with a local initialization converges linearly for smooth activations with bounded second derivatives for the regression problem, \cite{zhang2018learning} showed that gradient descent with tensor initialization converges linearly to a neighborhood of the ground truth using ReLU activations, and \cite{li2018nonconvex} showed the linear convergence of gradient descent with the spectral initialization using quadratic activations. For CNN with ReLU activations, \cite{brutzkus2017globally} shows that gradient descent converges to the ground truth with random initialization for the population risk function based on the squared loss under Gaussian inputs. Moreover, \cite{du2017convolutional} shows that gradient descent successfully learns a two-layer convolutional neural network despite the existence of bad local minima. From a technical perspective, our study differs from all the aforementioned work in that the cross-entropy loss function we analyze has a very different form. Furthermore, we study the model recovery classification problem under the multi-neuron case, which has not been studied before.

Finally, we note that several papers study one-hidden-layer or two-layer neural networks with different structures under Gaussian input. For example, \cite{goel2018learning} studied the overlapping convolutional neural network, \cite{li2017convergence} studied a two-layer feedforward networks with ReLU activations and identity mapping, and \cite{feizi2017porcupine} introduced the Porcupine Neural Network. 

\subsection{Paper Organization and Notations}

The rest of the paper is organized as follows. Section~\ref{sec:results} presents the main results on local geometry and local linear convergence of gradient descent. Section~\ref{sec:initialization} discusses the initialization based on the tensor method. 
Numerical examples are demonstrated in Section~\ref{sec:numerical}, and finally, conclusions are drawn in Section~\ref{sec:conclusions}. Details of the technical proofs are delayed in the supplemental materials.

Throughout this paper, we use boldface letters to denote vectors and matrices, e.g. $\bw$ and $\bW$. The transpose of $\bW$ is denoted by $\bW^{\top}$, and $\| \bW\|$, $\|\bW\|_{\mathrm{F}}$ denote the spectral norm and the Frobenius norm. For a positive semidefinite (PSD) matrix $\bA$, we write $\bA \succeq 0$. The identity matrix is denoted by $\bI$. The gradient and the Hessian of a function $f(\bW)$ is denoted by $\nabla f(\bW)$ and $\nabla^2 f(\bW)$, respectively. 

Denote $\|\cdot\|_{\psi_{1}}$ as the sub-exponential norm of a random variable.  We use $c,C,C_1,\ldots$ to denote constants whose values may vary from place to place. For nonnegative functions $f (x  ) $ and $g  (x  )$, $f (x  )= O\left(g  (x  ) \right) $ means there exist positive constants $c$ and $a$ such that $f (x  )\leq c g (x  )  $ for all $x\geq a$; $f (x  )= \Omega\left(g  (x  ) \right) $ means there exist positive constants $c$ and $a$ such that $f (x  )\geq c g (x  )  $ for all $x\geq a$. 

\section{Gradient Descent and its Performance Guarantee} \label{sec:results}

To estimate the network parameter $\bW^{\star}$, since~\eqref{Eq:Object} is a highly nonconvex function, vanilla gradient descent with an arbitrary initialization may get stuck at local minima. Therefore, we implement gradient descent (GD) with a well-designed initialization scheme that is described in details in Section~\ref{sec:initialization}. In this section, we focus on the performance of the local update rule 
$$\bW_{t+1} = \bW_{t} - \eta \nabla f_{n}\left(\bW_{t} \right),  $$ 
where $\eta$ is the constant step size. The algorithm is summarized in Algorithm~\ref{al:Global_Convergent}.
            
 \begin{algorithm}
 	\caption{Gradient Descent (GD)}\label{al:Global_Convergent}
 	\textbf{Input}: Training data $\left\lbrace\left(\bx_{i},y_{i} \right)  \right\rbrace_{i=1}^{n} $, step size $\eta$, iteration $T$ \\
 	\textbf{Initialization}: $\bW_{0} \gets \textsc{Initialization}\left(\left\lbrace\left(\bx_{i},y_{i} \right)  \right\rbrace_{i=1}^{n}  \right) $ \\
 	\textbf{Gradient Descent}: for $t=0,1,\cdots,T-1$ 
	 $$\bW_{t+1} = \bW_{t} - \eta \nabla f_{n}\left(\bW_{t} \right) . $$ 
	 \textbf{Output}: $\bW_{T}$
 \end{algorithm}


Note that throughout the execution of GD, the same set of training samples is used which is the standard implementation of gradient descent. Consequently the analysis is challenging due to the statistical dependence of the iterates with the data.

\subsection{Uniform local strong convexity}
We first characterize the local strong convexity of $f_n(\cdot)$ in a neighborhood of the ground truth.  We use the Euclidean ball to denote the local neighborhood of $\bW^{\star}$ for FCN or of $\bw^{\star}$ for CNN. 
\begin{subequations}
\begin{align}
\mathbb{B}\left(\bW^{\star},r   \right) &=\left\{ \bW\in\mathbb{R}^{d\times K}: \|\bW-\bW^{\star}\|_{\mathrm{F}} \leq r   \right\}, \\
 \mathbb{B}\left( \bw^{\star},r \right) &= \left\lbrace\bw\in \mathbb{R}^{m}:\|\bw-\bw^{\star}\|_{2}\leq r   \right\rbrace,
 \end{align}
\end{subequations}
where $r$ is the radius of the ball. With slight abuse of notations, we will drop the subscript FCN or CNN for simplicity, whenever it is clear from the context that the result is for FCN when the argument is $\bW\in\mathbb{R}^{d\times K}$ and for CNN when the argument is $\bw\in\mathbb{R}^m$. Further, $\sigma_{i}\left(\bW \right)$ denotes the $i$-th largest singular value of $\bW^{\star}$. Let the condition number be $\kappa = \sigma_1/\sigma_{K}$, and $\lambda =  \prod_{i=1}^{K} \left(\sigma_{i} /\sigma_{K} \right)$. Moreover, we introduce an important quantity $\rho\left(\sigma \right) $ regarding $\phi(z)$, the sigmoid activation function, that captures the geometric properties of the loss function for neural networks \eqref{eq:FNN:Classification_model} and \eqref{eq:CNN:Classification_model}.
		\begin{definition}[Key quantity for FCN]\label{def:rho_FNN} 
			Let $z\sim\mathcal{N}\left(0,1 \right) $ and define $\alpha_q(\sigma) = {\mathbb{E}}[\phi'(\sigma \cdot z) z^q], \forall q\in \{0,1,2\}$, and $\beta_q(\sigma) = {\mathbb{E}} [\phi'(\sigma \cdot z)^2 z^q ] , \forall q\in \{0,2\}.$ Define $\rho_{\mathrm{FCN}}(\sigma)$ as
			\begin{align*} 
			\rho_{\mathrm{FCN}}(\sigma)  = \min \left\{  \beta_0(\sigma) -\alpha_0^2(\sigma) , 
			\beta_2(\sigma) - \alpha_2^2(\sigma) \right\}  - \alpha_1^2(\sigma).
			\end{align*}
		\end{definition}
		
		\begin{definition}[Key quantity for CNN]\label{def:rho_CNN} 
			Let $z\sim\mathcal{N}\left(0,\sigma^{2} \right) $ and define $\rho_{\mathrm{CNN}}(\sigma)$ as
			\begin{align*}
			\rho_{\mathrm{CNN}}(\sigma) = \min\left\lbrace  \mathbb{E} [\left( \phi^{\prime} (z  )z\right)^{2}    ], \mathbb{E} [  \phi'\left(z \right)^2   ]  \right\rbrace .
			\end{align*} 
		\end{definition}
Note that Definition~\ref{def:rho_FNN} for FCN is different from that in \cite[Property 3.2]{zhong17a} but consistent with \cite[Lemma D.4]{zhong17a} which removes the third term in \cite[Property 3.2]{zhong17a}. For the activation function considered in this paper, the first two terms suffice. Definition~\ref{def:rho_CNN} for CNN is a newly distilled quantity in this paper tailored to the special structure of CNN.

The quantity $\rho\left(\sigma \right) $ plays an important role in the following theorem which guarantees the Hessian of the empirical risk function in the local neighborhood of the ground truth is positive definite with high probability for both FCN and CNN. 

\begin{theorem}[Local Strong Convexity]\label{PD_Hessian_Classification}
Consider the classification model with FCN \eqref{eq:FNN:Classification_model} or CNN \eqref{eq:CNN:Classification_model} and the sigmoid activation function.
\begin{itemize}
\item For FCN, assume $\|\bw^{\star}_{k}\|_{2}\leq 1$ for all $k$. There exist constants $c_1$ and $c_2$ such that as soon as sample size
$$ n_{\mathrm{FCN}} \geq c_{1}\cdot  dK^{5} \log^2 d \cdot \left(\frac{\kappa^{2} \lambda}{\rho_{\mathrm{FCN}}\left(\sigma_{K} \right) } \right)^{2},$$
with probability at least $1-d^{-10}$,
we have for all $ \bW\in\mathbb{B}(\bW^{\star},r_{\mathrm{FCN}})$,
\begin{align*}
 \Omega\left(\frac{1}{K^{2}}\cdot \frac{\rho_{\mathrm{FCN}}\left(\sigma_K \right) }{\kappa^2 \lambda} \right) \cdot \bI \preceq \nabla^{2}f_{n} \left(\bW  \right)   \preceq  \Omega(1) \cdot \bI  ,  
\end{align*} 
where $r_{\mathrm{FCN}} :=    \frac{c_{2}}{\sqrt{K}}\cdot \frac{\rho_{\mathrm{FCN}}\left(\sigma_{K} \right) }{\kappa^{2}\lambda} $.

\item  For CNN, assume $\|\bw^{\star}\|_{2}\leq 1$. There exist constants $c_3$ and $c_4$ such that as soon as sample size
$$ n_{\mathrm{CNN}} \geq c_{3}\cdot  dK^{5} \log^2 d \cdot \left(  \frac{1}{\rho_{\mathrm{CNN}}\left(\|\bw^{\star}\|_{2} \right) }\right)^{2},  $$
with probability at least $1-d^{-10}$,
we have for all $ \bw\in\mathbb{B}(\bw^{\star},r_{\mathrm{CNN}})$,
\begin{align*}
\Omega\left( \frac{1}{K}\cdot\rho_{\mathrm{CNN}}\left(\|\bw^{\star}\|_{2} \right)  \right) \cdot \bI \preceq \nabla^{2}f_{n} \left(\bw  \right)   \preceq \Omega( K) \cdot \bI  , 
\end{align*} 
where $r_{\mathrm{CNN}} :=   \frac{c_4}{K^{2}}\cdot \rho_{\mathrm{CNN}}\left(\|\bw^{\star}\|_{2} \right) $.

\end{itemize}
 
\end{theorem}
We note that for FCN \eqref{eq:FNN:Classification_model}, all column permutations of $\bW^{\star}$ are equivalent global minimum of the loss function, and Theorem~\ref{PD_Hessian_Classification} applies to all such permutation matrices of $\bW^{\star}$. 
The proof of Theorem \ref{PD_Hessian_Classification} is outlined in Appendix~\ref{append:Proof of strong convexity}. 

\begin{figure}[t]
	\centering
	\includegraphics[width=0.4\textwidth]{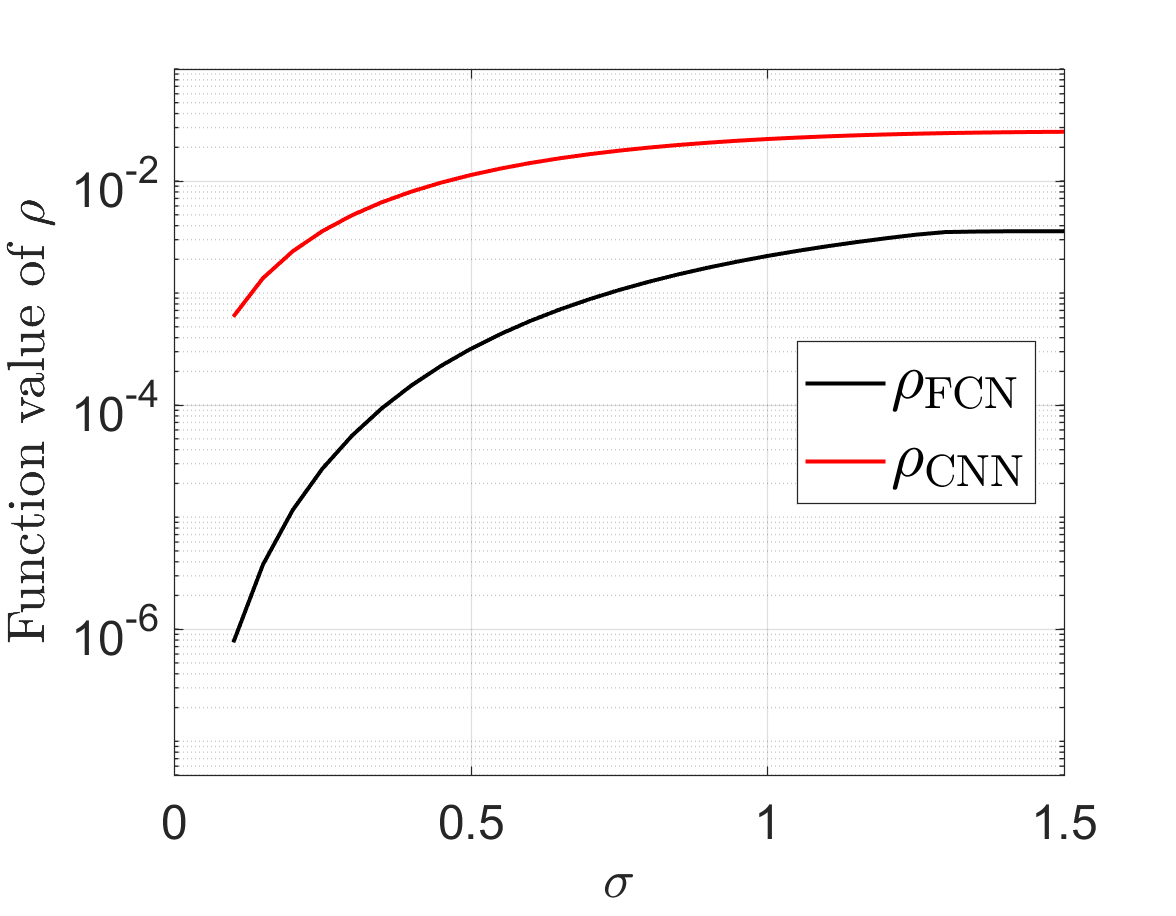}
	\caption{Illustration $\rho\left(\sigma \right) $ for both FCN and CNN with the sigmoid activation.}\label{fig:rho_sigmoid}	
\end{figure}

A pivot observation from the lower bound of the Hessian is that the sign of $\rho\left(\cdot \right) $ will determine whether the Hessian is positive definite or not, since $K, \kappa, \lambda$ are all positive. We depict $\rho(\sigma)$ as a function of $\sigma$ in a certain range for the sigmoid activation in Fig.~\ref{fig:rho_sigmoid}. It can be seen from Fig.~\ref{fig:rho_sigmoid} that $\rho\left(\sigma \right) $ is monotonic increasing when $\sigma$ increases, and we have $\rho(\sigma)>0$ as long as $\sigma> 0$. When $\bW^{\star}$ is orthogonal, $\kappa$ and $\lambda$ are both $1$, $\rho\left(\sigma \right) $ is a constant, hence the lower bound of Hessian is on the order of $\frac{1}{K^{2}}$ for FCN. However, in the worst case where the columns of $\bW^{\star}$ is linear dependent, then $\kappa$, $\lambda$, $\rho\left(\sigma \right) $ are infinite, and the local strong convexity doesn't hold for FCN case. Furthermore, the value of $\rho_{\mathrm{CNN}}(\sigma)$ is much larger than $\rho_{\mathrm{FCN}}(\sigma)$ for the same input.  

Theorem \ref{PD_Hessian_Classification} guarantees that for both FCN \eqref{eq:FNN:Classification_model} and CNN \eqref{eq:CNN:Classification_model} the Hessian of the empirical cross-entropy loss function $f_n(\bW)$ is positive definite in a neighborhood of the ground truth $\bW^{\star}$, as long as the sample size $n$ is sufficiently large and the columns of $\bW^{\star}$ are linearly independent. 
The bounds in Theorem~\ref{PD_Hessian_Classification} depend on the dimension parameters of the network ($n$ and $K$), as well as the ground truth ($\rho_{\mathrm{FCN}}(\sigma_K)$, $\lambda$, $\rho_{\mathrm{CNN}}\left(\|\bw^{\star}\|_{2} \right)$).

\subsection{Performance Guarantees of GD}
For the classification problem, due to the nature of quantized labels, $\bW^{\star}$ is no longer a critical point of $f_n(\bW)$. By the strong convexity of the empirical risk function $f_n(\bW)$ in the local neighborhood of $\bW^{\star}$, there can exist at most one critical point in $\mathbb{B}(\bW^{\star},r)$, which is the unique local minimizer in $\mathbb{B}\left(\bW^{\star},r \right) $ if it exists. The following theorem shows that there indeed exists such a critical point $\widehat{\bW}_n$, which is provably close to the ground truth $\bW^{\star}$, and gradient descent converges linearly to $\widehat{\bW}_n$.  

\begin{theorem}[Performance Guarantees of Gradient Descent]\label{theorem:GD_convergence_FNN}
Assume the assumptions in Theorem~\ref{PD_Hessian_Classification} hold. Under the event that local strong convexity holds, 
\begin{itemize}
\item for FCN, there exists a critical point in $\mathbb{B}(\bW^{\star},r_{\mathrm{FCN}})$ such that
$$ \left\|\widehat{\bW}_n - \bW^{\star} \right\|_{\mathrm{F}} \leq c_1 \frac{K^{9/4} \kappa^{2} \lambda }{\rho_{\mathrm{FNN}}\left(\sigma_{K} \right) }  \sqrt{\frac{d \log n}{n}}  , $$
and if  the initial point $\bW_{0}\in \mathbb{B} (\bW^{\star},r_{\mathrm{FCN}})$, GD converges linearly to $\widehat{\bW}_n$, i.e.		
	\begin{align*}
\hspace{-0.1in}	\left\|\bW_{t}-\widehat{\bW}_n \right\|_{\mathrm{F}} \leq \left(1-  \frac{c_2\eta \rho_{\mathrm{FCN}}\left(\sigma_K \right) }{K^2\kappa^2 \lambda} \right)^{t}  \left\|\bW_{0}-\widehat{\bW}_n \right\|_{\mathrm{F}},
	\end{align*} 
for $\eta \leq c_3$, where $c_1,c_2,c_3$ are constants;

\item for CNN, there exists a critical point in $\mathbb{B}(\bw^{\star},r_{\mathrm{CNN}})$ such that
$$ \left\|\widehat{\bw}_n - \bw^{\star} \right\|_{2} \leq c_{4} \frac{K}{\rho_{\mathrm{CNN}}\left(\|\bw^{\star}\|_{2} \right) }\cdot \sqrt{\frac{d\log n}{n}} , $$
and if  the initial point $\bw_{0}\in \mathbb{B} (\bw^{\star},r_{\mathrm{CNN}})$, GD converges linearly to $\widehat{\bw}_n$, i.e.		
	\begin{align*}
 \left\|\bw_{t}-\widehat{\bw}_n \right\|_{2} \leq \left(1-\frac{c_5 \eta \rho_{\mathrm{CNN}}\left(\|\bw^{\star}\|_{2} \right) }{K}  \right)^{t}  \left\|\bw_{0}-\widehat{\bw}_n \right\|_{2},
	\end{align*} 
for $\eta \leq c_6/K $, where $c_4,c_5,c_6$ are constants.
\end{itemize}
 \end{theorem}

Similarly to Theorem~\ref{PD_Hessian_Classification}, for FCN \eqref{eq:FNN:Classification_model} Theorem~\ref{theorem:GD_convergence_FNN} also holds for all column permutations of $\bW^{\star}$. 
The proof can be found in Appendix~\ref{proof: performance guarantee gradient descent}. Theorem~\ref{theorem:GD_convergence_FNN} guarantees that the existence of critical points in the local neighborhood of the ground truth, which GD converges to, and also shows that the critical points converge to the ground truth $\bW^{\star}$ at the rate of $O(K^{9/4}\sqrt{d\log n/n})$ for FCN \eqref{eq:FNN:Classification_model} and $O\left(K \sqrt{d\log n/n}\right) $ for CNN\eqref{eq:CNN:Classification_model} with respect to increasing the sample size $n$. Therefore, $\bW^{\star}$ can be recovered consistently as $n$ goes to infinity. Moreover, for both FCN \eqref{eq:FNN:Classification_model} and CNN \eqref{eq:CNN:Classification_model} gradient descent converges linearly to $\hat{\bW}_n$ (or resp. $\hat{\bw}_n$) at a linear rate, as long as it is initialized in the basin of attraction. To achieve $\epsilon$-accuracy, i.e. $\left\|\bW_{t}-\widehat{\bW}_n \right\|_{\mathrm{F}} \leq \epsilon$ (or resp. $\left\|\bw_{t}-\widehat{\bw}_n \right\|_{2} \leq \epsilon$), it requires a computational complexity of $O\left(ndK^4\log\left( 1/\epsilon\right)   \right) $ (or resp. $O\left(ndK^2\log\left( 1/\epsilon\right)   \right) $), which is linear in $n$, $d$ and $\log(1/\epsilon)$.

\section{Initialization via Tensor Method} \label{sec:initialization}
Our initialization adopts the tensor method proposed in \cite{zhong17a}. The initialization method works for the FCN model directly, and works for the CNN model with slight modification as presented in \cite{zhong2017learning}. To avoid unnecessary repetitions from the previous work, we focus on the FCN case to outline the algorithm and remark the difference. We recommend the readers refer to \cite{zhong17a, zhong2017learning} for more details.
 
\subsection{Preliminary and Algorithm} 
We start with introducing the necessary definitions which can be found in \cite{zhong17a}. We first define a product $\widetilde{\otimes}$ as follows. If $ \bv \in \mathbb{R}^d$ is a vector and $\bI$ is the identity matrix, then
$ \bv \widetilde{\otimes} \bI = \sum_{j=1}^d [ \bv\otimes \be_j\otimes \be_j+ \be_j\otimes \bv\otimes \be_j+ \be_j \otimes \be_j \otimes \bv] $. If $\bM$ is a symmetric rank-$r$ matrix factorized as $\bM = \sum_{i=1}^r \bs_i \bv_i \bv_i^\top$ and $\bI$ is the identity matrix, then
\begin{align}\label{def:outer product}
	\bM \widetilde{\otimes} \bI = \sum_{i=1}^r \bs_i \sum_{j=1}^d \sum_{l=1}^6 \bA_{l,i,j},
\end{align}
where $\bA_{1,i,j} = \bv_i \otimes  \bv_i \otimes \be_j \otimes \be_j$, $\bA_{2,i,j} = \bv_i \otimes \be_j \otimes \bv_i \otimes \be_j$, $\bA_{3,i,j}= \be_j \otimes \bv_i \otimes \bv_i \otimes \be_j $, $\bA_{4,i,j}= \bv_i \otimes \be_j \otimes \be_j  \otimes  \bv_i$, $\bA_{5,i,j}= \be_j \otimes \bv_i \otimes \be_j  \otimes \bv_i$ and $\bA_{6,i,j} = \be_j \otimes \be_j  \otimes \bv_i \otimes  \bv_i $.
This allows us to introduce the following quantities.
\begin{definition}\label{def:M_m}
Define $\bM_1$, $\bM_2$, $\bM_3$, $\bM_4$ and $m_{1,i}$, $m_{2,i}$, $m_{3,i}$, $m_{4,i}$ as follows: \\
	$\bM_1 = \mathbb{E} [y \cdot \bx]$,\\
	$\bM_2=  \mathbb{E} [y \cdot (\bx\otimes \bx -\bI)]$, \\
	$\bM_3 = \mathbb{E} [y \cdot ( \bx^{\otimes 3} - \bx \widetilde{\otimes} \bI)]$,\\
	$\bM_4 = \mathbb{E} [y \cdot ( \bx^{\otimes 4} -  (\bx\otimes \bx) \widetilde{\otimes} \bI +   \bI \tilde \otimes \bI)]$,\\
	$m_{l,i} = g_{l,i}\left(\|\bw_i^{\star}\| \right), \forall l=0,1,2,3,4,$\\
	where $g_{1,i}\left(\sigma \right)  = \gamma_1(\sigma) $, 
	$g_{2,i}\left(\sigma \right) = \gamma_2(\sigma) -\gamma_0(\sigma)$,
	$g_{3,i}\left(\sigma \right) = \gamma_3 ( \sigma)- 3\gamma_1( \sigma)$,
	$g_{4,i}\left(\sigma \right) =  \gamma_4(\sigma)+3\gamma_0(\sigma) - 6\gamma_2(\sigma)$,	and $\gamma_j(\sigma) = \mathbb{E}_{z\sim \mathcal{N}(0,1)}[ \phi(\sigma\cdot z)z^j], \; \forall j=0,1,2,3,4.$ 
\end{definition}

We further define a tensor operation as follows. For a tensor $\bT\in \mathbb{R}^{n_1\times n_2 \times n_3}$ and three matrices $\bA\in \mathbb{R}^{n_1 \times d_1}, \bB\in \mathbb{R}^{n_2 \times d_2}, \bC\in \mathbb{R}^{n_3 \times d_3} $, the $\left(i,j,k \right) $-th entry of the tensor $\bT\left(\bA,\bB,\bC \right) $ is given by
\begin{equation}\label{tensor_product}
\sum_{i^{\prime}}^{n_1} \sum_{j^{\prime}}^{n_2} \sum_{k^{\prime}}^{n_3}
\bT_{i^{\prime},j^{\prime},k^{\prime}} \bA_{i^{\prime},i} \bB_{j^{\prime},j} \bC_{k^{\prime},k}.
\end{equation}  
Armed this with definition, we define the following useful quantities.
\begin{definition}\label{def:P2_P3}
	Let $\balpha \in \mathbb{R}^{d}$ denote a randomly picked vector. We define $\bP_2$ and $\bP_3$ as follows: $\bP_{2} = \bM_{j_2}(\bI,\bI,  \balpha,\cdots, \balpha) $, where $j_2 = \min \{j\geq 2| \bM_{j}\neq 0\}$, and
	$\bP_{3} =  \bM_{j_3}(\bI,\bI,\bI,\balpha,\cdots, \balpha) $, 
	where $j_3 =  \min \{j\geq 3| \bM_{j}\neq 0\} $.
\end{definition}
We further denote $\overline{ \bw} = \bw/\| \bw\|$. An important implication of Definition \ref{def:M_m} and \ref{def:P2_P3} is that the non-zero matrix $P_{2}$ and non-zero tensor $P_{3}$ is in the form of $\sum_{i=1}^{K}m_{j_{2},i}\left(\alpha^{\top}\overline{\bw}_{i}^{\star} \right)^{j_{2}-2}\overline{\bw}_{i}^{\star\otimes 2} $, $\sum_{i=1}^{K}m_{j_{3},i}\left(\alpha^{\top}\overline{\bw}_{i}^{\star} \right)^{j_{3}-3}\overline{\bw}_{i}^{\star\otimes 3} $, see\cite[Claim 5.5]{zhong17a}. The basic strategy is to extract the direction, magnitude information from the empirical version of $\bP_{2}$ and $\bP_{3}$. Hence estimating $\bW^{\star}$ can be decomposed as the following two steps. 
\begin{itemize}
	\item[Step 1] Estimate the direction of each column of $\bW^{\star}$ by decomposing $\bP_{2}$ to approximate the subspace spanned by $\left\lbrace \overline{\bw}_1^{\star}, \overline{\bw}_2^{\star}, \cdots, \overline{\bw}_K^{\star} \right\rbrace$ (denoted by $\bV$), then reduce the third-order tensor $P_{3}$ to a lower-dimension tensor $\bR_{3} = \bP_{3}\left(\bV, \bV, \bV \right)\in \mathbb{R}^{K\times K \times K}$, and apply non-orthogonal tensor decomposition on $\bR_{3}$ to output the estimate $s_{i}\bV^{\top}\overline{ \bw}_i^{\star}$, where $s_{i}\in \left\lbrace 1,-1 \right\rbrace $ is a random sign. 
	\item[Step 2] Approximate the magnitude of $\bw_{i}^{\star}$ and the sign $s_{i}$ by solving a linear system of equations.
\end{itemize}

The initialization algorithm based on the tensor method is outlined in Algorithm \ref{algorithm:initial}. For more implementation details about Algorithm \ref{algorithm:initial}, e.g., power method, we refer to \cite{zhong17a}. 
\begin{algorithm}[ht]
	\caption{Initialization via Tensor Method}	\label{algorithm:initial} 
	\begin{algorithmic}[1]  
		\REQUIRE ~Partition $n$ pairs of data $\left\lbrace \left(\bx_{i},y_{i} \right)  \right\rbrace_{i=1}^{n} $ into three subsets $\mathcal{D}_{1},\mathcal{D}_{2},\mathcal{D}_{3}$. \\
		\ENSURE
		 \STATE Estimate $\widehat{\bP}_{2}$ of $\bP_{2}$ from data set $\mathcal{D}_{1}$. \\
		 \STATE $\bV\leftarrow \textsc{PowerMethod}(\widehat{\bP}_2,K)$.\\
		 \STATE  Estimate $\widehat{\bR}_{3}$ of $\bP_{3}(\bV,\bV,\bV)$ from data set $\mathcal{D}_{2}$. \\
		 \STATE  $ \{ \widehat{\bu}_i\}_{i\in[K]}  \leftarrow \textsc{KCL}(\widehat{\bR}_3)$. \\
		 \STATE  $\{ \bw_i^{(0)}\}_{i\in[K]} \leftarrow $ \textsc{RecMag}$(\bV,\{ \widehat{\bu}_i\}_{i\in[K]}, \mathcal{D}_{3} )$.
	\end{algorithmic}
\end{algorithm}

\subsection{Performance Guarantee of Initialization}
For the classification problem, we make the following technical assumptions, similarly to \cite[Assumption 5.3]{zhong17a} for the regression problem.

\begin{assumption}\label{assumption:initial}
	The activation function $\phi(z)$ satisfies the following conditions:
	\begin{enumerate}
	\item If $M_j\neq 0 $, then 
	\begin{align*}
	\hspace{-0.25in} & \sum_{i=1}^{K}m_{j,i}\left({\bw_i^{\star}}^{\top}\balpha \right)^{j-2} \overline{\bw_{i}}^{\star} \overline{\bw_{i}}^{\star\top} \neq \bm{0}, \nonumber \\
	\hspace{-0.25in}  & \sum\limits_{i=1}^{K} m_{j,i}\left(\overline \bw_i^{\star\top}  \balpha \right)^{j-3} (\bV^\top \overline \bw_i^{\star}) \mathrm{vec}((\bV^\top \overline \bw_i^{\star})(\bV^\top \overline \bw_i^{\star})^\top)^\top  \neq 0, 
	\end{align*} 
	for $j\geq 3$.
	\item At least one of $M_3$ and $M_4$ is non-zero.	
	\end{enumerate}
\end{assumption}
Assumption~\ref{assumption:initial} is to guarantee that the key terms still contain the magnitude information about $\bw^{\star}_{j}$. It can be verified that for sigmoid activation $m_{3,i}$ is non-zero for $\sigma>0$, hence it will satisfy Assumption~\ref{assumption:initial}. 
Furthermore, we do not require the homogeneous assumption (i.e., $\phi(az)=a^pz$ for an integer $p$) required in \cite{zhong17a}, which can be restrictive. Instead, we assume the following condition on the curvature of the activation function around the ground truth, which holds for a larger class of activation functions such as sigmoid and tanh.

\begin{assumption}\label{assumption:inverse}
Let $l_1$ be the index of the first nonzero $M_i$ where $i=1,\ldots,4$. For the activation function $\phi\left(\cdot \right) $, 
there exists a positive constant $\delta$ such that $g_{l_1,i}(\cdot)$ is strictly monotone over the interval $\left(\|\bw_{i}^{\star}\|-\delta,\|\bw_{i}^{\star}\|+\delta  \right) $,
and the derivative of $g_{l_1,i}(\cdot)$ is lower bounded by some constant for all $i$. 
\end{assumption}
It can be numerically verified that sigmoid activation will also satisfy Assumption~\ref{assumption:inverse}. We next present the performance guarantee for the initialization algorithm in the following theorem.
\begin{theorem}\label{theorem:initial_guarantee}
For the classification model \eqref{eq:FNN:Classification_model}, under Assumptions \ref{assumption:initial} and \ref{assumption:inverse}, for any $0<\epsilon<1$ and $\zeta>1$, if the sample size $n \geq d\cdot\mathrm{poly}\left( K,\kappa,\zeta,\log d,1/\epsilon \right)$, then the output $\bW_{0}\in \mathbb{R}^{d\times K}$ of Algorithm \ref{algorithm:initial} satisfies  
	\begin{equation}
	\|\bW_{0} - \bW^{\star}\|_{\mathrm{F}} \leq \epsilon \mathrm{poly}\left(K,\kappa \right) \|\bW^{\star}\|_{\mathrm{F}}, 
	\end{equation} 
	with probability at least $1-d^{-\Omega\left( \zeta\right) }$.
\end{theorem}
The proof of Theorem \ref{theorem:initial_guarantee} consists of (a) showing the estimation of the direction of $\bW^{\star}$ is sufficiently accurate and (b) showing the approximation of the norm of $\bW^{\star}$ is accurate enough. The proof of part (a) is the same as that in \cite{zhong17a}, but our argument in part (b) is different, where we relax the homogeneous assumption on activation functions. More details can be found in the supplementary materials in Appendix~\ref{Appendix:initial}.



\section{Numerical Experiments} \label{sec:numerical} 
\begin{figure*}[ht]
\begin{center}
	\begin{tabular}{cc}
		\includegraphics[width=0.4\textwidth]{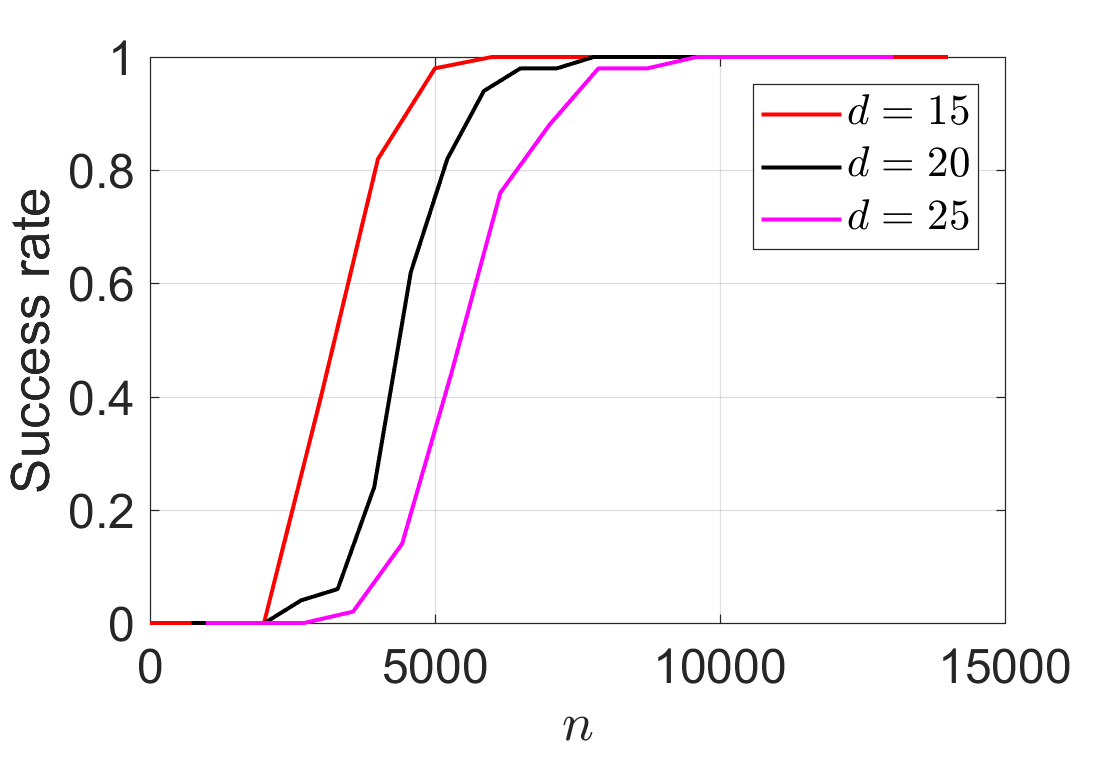} &\includegraphics[width=0.4\textwidth]{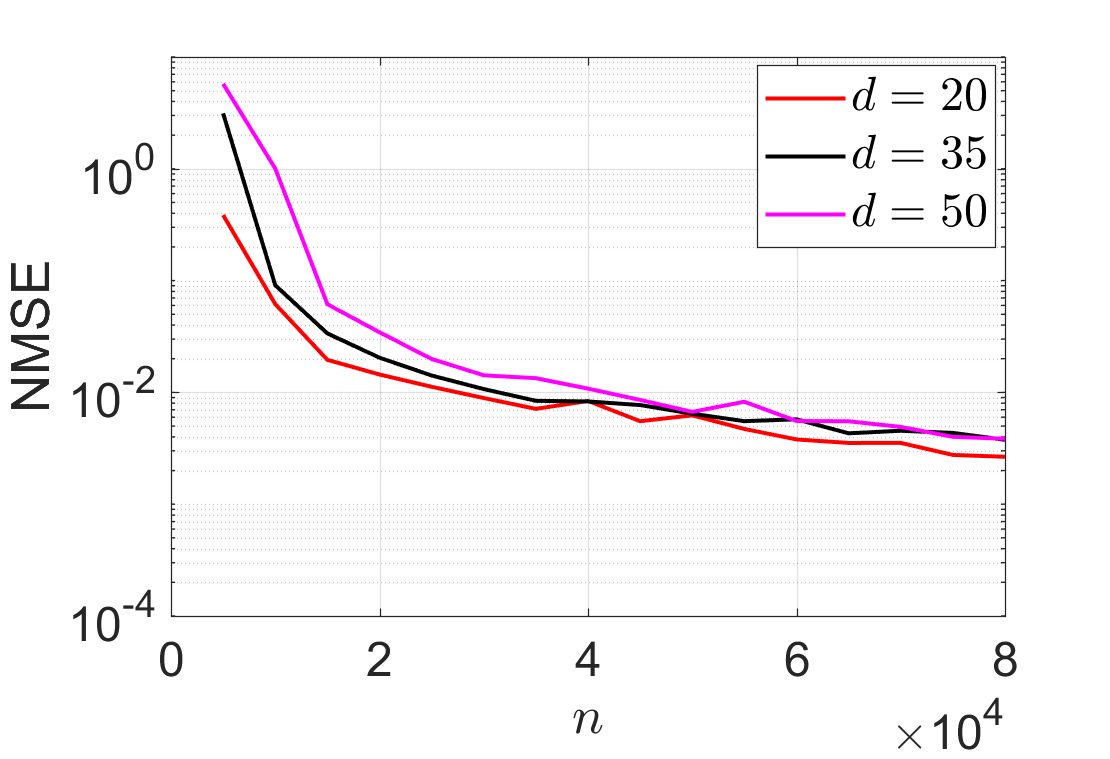} \\
		(a)  & (b) 
	\end{tabular}
\end{center}
	\caption{For FCN \eqref{eq:FNN:Classification_model} fix $K=3$. (a) Success rate of converging to the same local minima with respect to the sample complexity for various $d$ with threshold $10^{-4}$; (b) Average estimation error of gradient descent in a local neighborhood of the ground truth with respect to the sample complexity for various $d$. The x-axis is scaled to illuminate the correct scaling between $n$ and $d$.}  
	\label{fig:Numerical}
\end{figure*}

For FCN, we first implement gradient descent to verify that the empirical risk function is strongly convex in the local region around $\bW^{\star}$. If we initialize multiple times in such a local region, it is expected that gradient descent converges to the same critical point $\widehat{\bW}_n$, with the same set of training samples. Given a set of training samples, we randomly initialize multiple times, and then calculate the variance of the output of gradient descent. Denote the output of the $\ell$th run as $\hat{\bw}_n^{(\ell)}=\mathrm{vec}(\hat{\bW}_n^{(\ell)})$ and the mean of the runs as $\bar{\bw}$. The error is calculated as
$\mathrm{SD}_n = \sqrt{\frac{1}{L} \sum_{\ell=1}^L \|\hat{\bw}_n^{(\ell)}-\bar{\bw} \|^2}$,
where $L=20$ is the total number of random initializations. Adopted from \cite{mei2016landscape}, it quantifies the standard deviation of the estimator $\widehat{\bW}_n$ under different initializations with the same set of training samples. We say an experiment is successful, if $\mathrm{SD}_{n}\leq 10^{-4}$. We generate the ground truth $\bW^{\star}$ from Gaussian matrices, and the training samples are generated using the FCN \eqref{eq:FNN:Classification_model}. \Cref{fig:Numerical}~(a) shows the successful rate of gradient descent by averaging over $50$ sets of training samples for each pair of $n$ and $d$, where $K=3$ and $d=15,20,25 $ respectively. The maximum iterations for gradient descent is set as $\mathrm{iter}_{\max}=3500$. It can be seen that as long as the sample complexity is large enough, gradient descent converges to the same local minima with high probability. 

\begin{figure*}[ht]
	\begin{center}
		\begin{tabular}{cc}
			\includegraphics[width=0.4\textwidth]{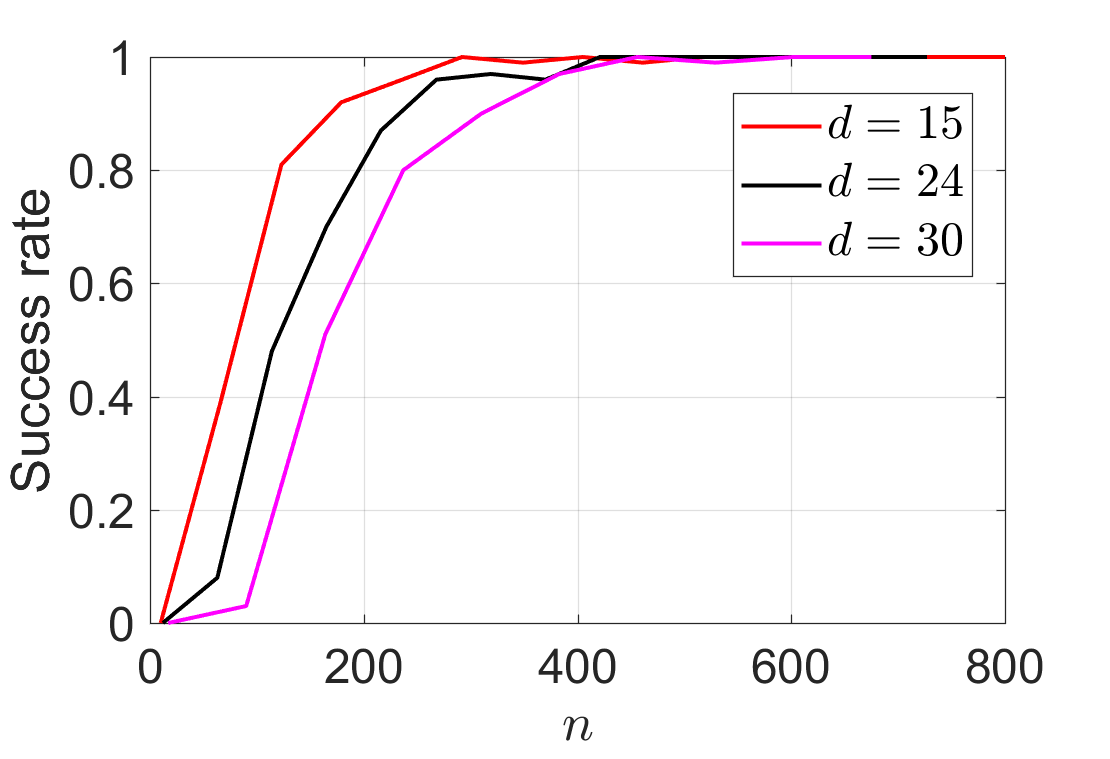} &\includegraphics[width=0.4\textwidth]{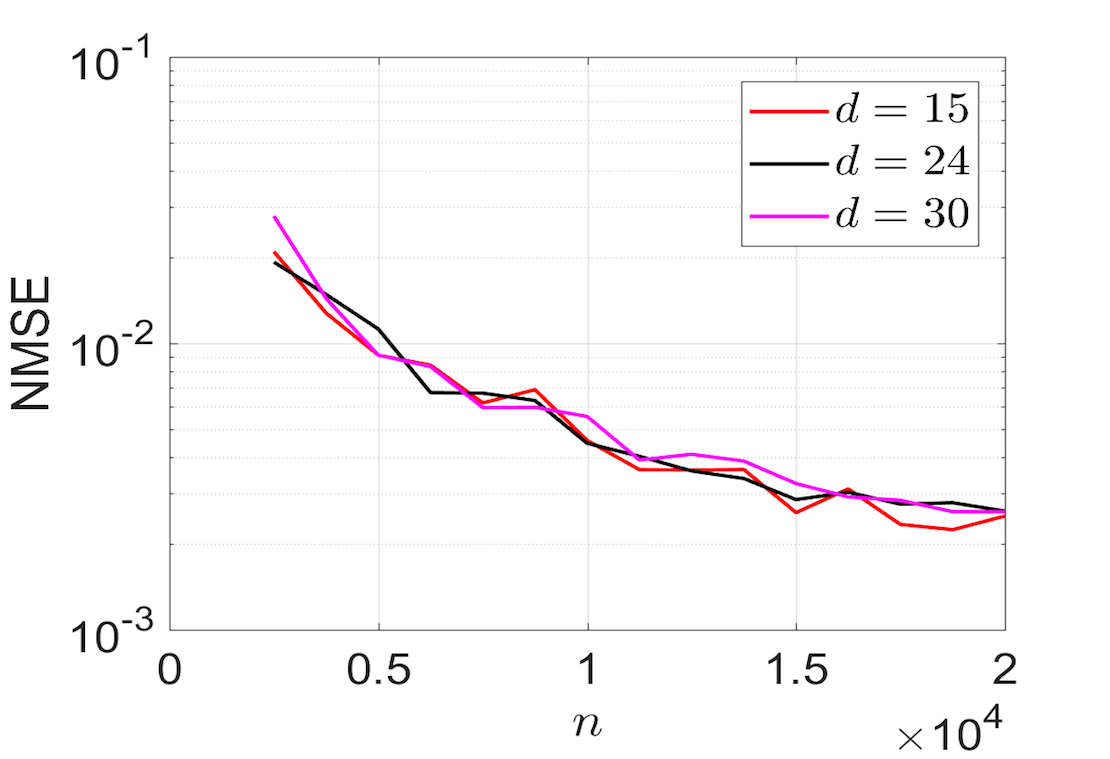} \\
			(a)  & (b) 
		\end{tabular}
	\end{center}
	\caption{For CNN \eqref{eq:CNN:Classification_model}, fix $K=3$. (a) Success rate of converging to the same local minima with respect to the sample complexity for various $d$ with threshold $10^{-14}$; (b) Average estimation error of gradient descent in a local neighborhood of the ground truth with respect to the sample complexity for various $d$. The x-axis is scaled to illuminate the correct scaling between $n$ and $d$.  }\label{Fig:experiment_of_CNN}
\end{figure*}
We next show that the statistical accuracy of the local minimizer for gradient descent if it is initialized close enough to the ground truth. Suppose we initialize around the ground truth such that $\|\bW_{0} - \bW^{\star}\|_{\mathrm{F}}\leq 0.1\cdot \|\bW^{\star}\|_{\mathrm{F}} $. We calculate the average estimation error as $\sum_{\ell=1}^L \|\widehat{\bW}_n^{(\ell)}-\bW^{\star}\|_{\mathrm{F}}^{2}/(L\|\bW^{\star}\|_{\mathrm{F}}^{2})$ over $L= 100$ Monte Carlo simulations with random initializations. Fig.~\ref{fig:Numerical}~(b) shows the average estimation error with respect to the sample complexity when $K=3$ and $d=20,35, 50$ respectively. It can be seen that the estimation error decreases gracefully as we increase the sample size and matches with the theoretical prediction of error rates reasonably well.

Similarly, for CNN, we first verify that the empirical risk function is locally strongly convex using the same method as before. We generate the entries of true weights $\bw^{\star}$ from standard Gaussian distribution, and generate the training samples using the CNN model \eqref{eq:CNN:Classification_model}. In Fig.~\ref{Fig:experiment_of_CNN}~(a), we say an experiment is successful if $\mathrm{SD}_{n}\leq 10^{-14}$, and the successful rate is calculated over $100$ sets of training samples with $K=3$ and $d=15,24,30$ respectively. Then we verify the performance of gradient descent in Fig.~\ref{Fig:experiment_of_CNN}~(b). Suppose we initialized in the neighborhood of $\bw^{\star}$, i.e., $\|\bw_{0}-\bw^{\star}\|_{2}\leq 0.9\cdot \|\bw^{\star}\|_{2}$, for fixed $d,K,n$, the average error is calculated over $L=100$ Monte Carlo simulations. It can be seen that the error decreases as we increase the number of samples.

%

\section{Conclusions} \label{sec:conclusions}

In this paper, we have studied the model recovery problem of a one-hidden-layer neural network using the cross-entropy loss in a multi-neuron classification problem. In particular, we have characterized the sample complexity to guarantee local strong convexity in a neighborhood (whose size we have characterized as well) of the ground truth when the training data are generated from a classification model for two types of neural network models: fully-connected network and non-overlapping convolutional network. This guarantees that with high probability, gradient descent converges linearly to the ground truth if initialized properly. In the future, it will be interesting to extend the analysis in this paper to more general class of activation functions, particularly ReLU-like activations.

\appendices
\section{Gradient and Hessian of Population Loss}
For the convenience of analysis, we first provide the gradient and the Hessian formula for the cross-entropy loss using FCN and CNN here. 

\subsection{The FCN case}
	Consider the population loss function $	f(\bW) = \mathbb{E} \left[ f_n (\bW) \right] = \mathbb{E}\left[ \ell \left(\bW;\bx \right)\right]$, where $\ell \left(\bW;\bx \right)$ is associated with network $ H_{\mathrm{FCN}}\left(\bW ,\bx \right) = \frac{1}{K} \sum_{k=1}^K \phi(\bw_k^{ \top}\bx)$. Hiding the dependence on $\bm{x}$ for notational simplicity, we can calculate the gradient and the Hessian as
	\begin{align}
	\mathbb{E}\left[\frac{\partial \ell\left(\bW \right)}{\partial \bw_{j}}\right] &=\mathbb{E}\left[- \frac{1}{K}  \frac{(y-H(\bW  ))}{H(\bW ) \left(1-H(\bW)  \right) }   \phi^{\prime}\left(\bw_{j}^{\top}\bx \right) \bx\right] ,
	\label{eq:gradient_formula} \\
	\mathbb{E}\left[\frac{\nabla^{2} \ell\left(\bW \right) }{\partial  \bw_{j}\partial \bw_{l} }\right] &=
	\mathbb{E}\left[ \xi_{ j,l }\left(\bW \right)    \cdot \bx\bx^{\top}  \right],\label{eq:hessian_formula}
	\end{align} 	
	for $1\leq j,l\leq K$. Here, when $j\neq l$, 
	\small
	\begin{align*}
	\xi_{j,l}\left(\bW  \right) = &\frac{1}{K^{2}} \phi^{\prime} \left(\bw_{j}^{\top}\bx \right)\phi^{\prime} \left(\bw_{l}^{\top}\bx \right)\cdot \frac{H\left(\bW \right) ^{2}+y-2y\cdot H\left(\bW \right)}{H^2  (\bW )\left(1-H  (\bW  ) \right)^2   } , 
	\end{align*}	
	\normalsize
	and when $j=l$, 
	\small
	\begin{align*}
	\xi_{j,j}\left(\bW \right) & =  \frac{1}{K^{2}} \phi^{\prime} \left(\bw_{j}^{\top}\bx \right)^{2}\cdot\frac{H\left(\bW \right) ^{2}+y-2y\cdot H\left(\bW \right)}{H^2  (\bW )\left(1-H  (\bW  ) \right)^2   } \\
	& \quad \quad  - \frac{1}{K} \phi^{\prime\prime} \left(\bw_{j}^{\top}\bx \right)\cdot 
	\frac{ y - H \left(\bW  \right)}{H  (\bW )\left(1-H  (\bW  ) \right)   }.
	\end{align*}
	\normalsize
	
\subsection{The CNN case} 
	For the CNN case, i.e., $H\left(\bw \right) := H_{\mathrm{CNN}}\left(\bw,\bx \right) = \frac{1}{K} \sum_{k=1}^K \phi(\bw^{\top}\bx^{\left(k \right) })$, 
	the corresponding gradient and Hessian of the population loss function $\ell(\bw)$ is given by
	\begin{align} 
	&\mathbb{E}\left[\frac{\partial \ell\left(\bw \right)}{\partial \bw}\right]
	=\mathbb{E}\left[- \phi^{\prime} (\bw^{\top}\bx^{ (1 ) }  )\cdot
	\frac{y-H\left(\bw \right)}{H\left(\bw \right)\left(1-H\left(\bw \right)  \right) } \cdot \bx^{\left(1\right) }\right], 
	\label{eq:gradient_formula_CNN}\\
	&\mathbb{E}\left[\frac{\nabla^{2} \ell\left(\bw \right) }{\partial  \bw^{2} }\right]= \mathbb{E}\left[  \sum_{j=1}^{K}\sum_{l=1}^{K}g_{j,l}\left(\bw \right) \bx^{\left( j\right) }\bx^{\left( l\right) \top}\right],    
	\label{eq:hessian_formula_CNN} 
	\end{align}	
where when $j\neq l$,    
\small
\begin{align*}
	g_{j,l}\left(\bw \right)=\frac{1}{K^{2}}\cdot \frac{H\left(\bw \right) ^{2}+y-2y\cdot H\left(\bw \right)}{\left(H\left(\bw \right)\left(1-H\left(\bw \right) \right)  \right)^{2} }  \phi^{\prime}\left(\bw^{\top}\bx^{\left( j\right) } \right)\phi^{\prime}\left(\bw^{\top}\bx^{\left( l\right) } \right),
\end{align*} 
\normalsize
and when $j=l $, 
	\small
\begin{align*}
	g_{j,j}\left(\bw \right) &= \frac{1}{K^{2}}\cdot\frac{H\left(\bw \right)^{2}+y-2y\cdot H\left(\bw \right)}{\left(H\left(\bw \right)\left(1-H\left(\bw \right) \right)  \right)^{2} } \cdot \phi^{\prime}\left(\bw^{\top}\bx^{\left( j\right) } \right)^{2} \\
	& - \frac{1}{K}\cdot\frac{y-H\left(\bw \right)}{H\left(\bw \right)\left(1-H\left(\bw \right) \right) }\cdot \phi^{\prime\prime}\left(\bw^{\top}\bx^{\left( j\right) } \right). 
\end{align*} 	
 \normalsize

\section{Proof of Theorem~\ref{PD_Hessian_Classification}}\label{append:Proof of strong convexity}
In order to show that the empirical loss possesses a local strong convexity, we follow the following steps:

\begin{enumerate}
	\item We first show that the Hessian $\nabla^2 f(\bW)$ of the population loss function is smooth with respect to $\nabla^2 f(\bW^{\star})$ (Lemma~\ref{lem:smooth_pop_local});
	\item We then show that $\nabla^2 f(\bW)$ satisfies local strong convexity and smoothness in a neighborhood of $\bW^{\star}$ with appropriately chosen radius, $\mathbb{B}(\bW^{\star},r)$, by leveraging similar properties of $\nabla^2 f(\bW^{\star})$ (Lemma~\ref{lem:Bounds_Hessian_pop_tru});
	\item Next, we show that  the Hessian of the empirical loss function $\nabla^2 f_n(\bW)$ is close to its population counterpart $\nabla^2 f(\bW)$ uniformly in $\mathbb{B}(\bW^{\star},r)$ with high probability (Lemma~\ref{Uniform_Convergence}).
	\item Finally, putting all the arguments together, we establish $\nabla^2 f_n(\bW)$ satisfies local strong convexity and smoothness in $\mathbb{B}(\bW^{\star},r)$.
\end{enumerate} 

To begin, we first show that the Hessian of the population risk is smooth enough around $\bW^{\star}$ in the following lemmas.
\begin{lemma}[Hessian Smoothness of Population Loss]\label{lem:smooth_pop_local}
	Suppose the loss $\ell\left(\cdot \right) $ associates with FCN \eqref{eq:FNN:Classification_model}, and assume $\|\bw^{\star}_{k}\|_{2}\leq 1$ for all $k$ and $\|\bW-\bW^{\star}\|_{\mathrm{F}} \leq 0.7$. Then we have
	\begin{equation}\label{Eq:Hessian smoothness FCN}
	\|\nabla^{2}f\left(\bW \right) - \nabla^{2}f\left(\bW^{\star} \right)   \| \leq \frac{C_{1}}{K^{\frac{3}{2}}} \cdot \|\bW-\bW^{\star}\|_{\mathrm{F}},  
	\end{equation}
	holds. Similarly, suppose the loss $\ell\left(\cdot \right) $ associates with CNN \eqref{eq:CNN:Classification_model}, and assume $\|\bw^{\star}\|_{2}\leq 1$ and  $\|\bw-\bw^{\star}\|_{2} \leq 0.7$. We have 
	\begin{align}\label{Eq:Hessian smoothness CNN}
	\|\nabla^{2}f\left(\bw \right) - \nabla^{2}f\left(\bw^{\star} \right)   \| \leq C_{2} \cdot K\cdot \|\bw-\bw^{\star}\|_{2}, 
	\end{align}
	holds. Here $C_{1}$ and $C_{2}$ denote some large constants.
\end{lemma} 
 The proof is provided in Appendix~\ref{proof: smoothness}. Together with the fact that $\nabla^2 f(\bW^{\star})$ be lower and upper bounded, Lemma~\ref{lem:smooth_pop_local} allows us to bound $\nabla^2 f(\bW)$ in a neighborhood around ground truth, given below.

\begin{lemma}[Local Strong Convexity and Smoothness of Population Loss]\label{lem:Bounds_Hessian_pop_tru}
	If the loss $\ell\left(\cdot \right) $ associates with FCN \eqref{eq:FNN:Classification_model}, there exists some constant $C_{1}$, such that
	\begin{align*}\label{Population_lowerbound}
	\frac{4}{K^{2}}\cdot \frac{\rho_{\mathrm{FCN}}\left(\sigma_{K} \right) }{\kappa^{2}\lambda} \cdot \bI \preceq \nabla^{2}f\left(\bW \right) \preceq C_{1}\cdot\bI, 
	\end{align*}
	holds for all $\bW\in\mathbb{B}(\bW^{\star},r_{\mathrm{FCN}})$ with $r_{\mathrm{FCN}}:=  \frac{C_{2}}{K^{\frac{1}{2}}}\cdot \frac{\rho_{\mathrm{FCN}}\left(\sigma_{K} \right) }{\kappa^{2}\lambda}$. Moreover, if loss $\ell\left(\cdot \right) $ associates with CNN \eqref{eq:CNN:Classification_model}, then we have \begin{align}
	C_{3}\cdot \frac{\rho_{\mathrm{CNN}}\left(\|\bw^{\star}\|_{2} \right) }{K}\cdot \bI \preceq \nabla^{2}f\left(\bw \right) \preceq C_{4}\cdot K\cdot \bI, 
	\end{align}
	holds for all $\bw\in \mathbb{B}\left(\bw^{\star},r_{\mathrm{CNN}} \right) $ with $r_{\mathrm{CNN}}:=C_{5}\cdot \frac{\rho_{\mathrm{CNN}}\left(\|\bw^{\star}\|_{2} \right)}{K^{2}} $. 
\end{lemma} 
The proof is provided in Appendix~\ref{proof: proof of local strong convexity}. The next step is to show the Hessian of the empirical loss function is close to the Hessian of the population loss function in a uniform sense, which can be summarized as follows. 

\begin{lemma}\label{Uniform_Convergence}
	If the loss $\ell\left(\cdot \right) $ associates with FCN \eqref{eq:FNN:Classification_model}, then there exists a constant $C$ such that as long as $n\geq C\cdot dK\log dK$, with probability at least $1-d^{-10}$, the following holds
	\begin{align}\label{Eq:FCN:uniform convergence}
	\underset{\bW\in \mathbb{B}\left(\bW^{\star},r_{\mathrm{FCN}} \right) }{\mathrm{sup}} \|\nabla^{2}f_{n}\left(\bW \right) - \nabla^{2}f\left(\bW \right)  \| \leq C \sqrt{\frac{dK \log n}{n}}, 
	\end{align}	
	where $r_{\mathrm{FCN}}:= \frac{C}{K^{\frac{1}{2}}}\cdot \frac{\rho\left(\sigma_{K} \right) }{\kappa^{2}\lambda} $. And if the loss $\ell\left(\cdot \right) $ associates with CNN \eqref{eq:CNN:Classification_model}, then we have 
	\begin{align}\label{Eq:CNN:uniform convergence}
	\underset{\bw\in \mathbb{B}\left(\bw^{\star},r_{\mathrm{CNN}} \right) }{\mathrm{sup}} \|\nabla^{2}f_{n}\left(\bw \right) - \nabla^{2}f\left(\bw \right)  \| \leq C K^{2}\sqrt{\frac{\frac{d}{K}\cdot \log\left(n \right) }{n}}, 
	\end{align}
	holds with probability at least $1-d^{-10}$, as long as $n\geq \frac{d}{K}\log\left(\frac{d}{K} \right) $, and   $r_{\mathrm{CNN}}:=C\cdot \frac{\rho_{\mathrm{CNN}}\left(\|\bw^{\star}\|_{2} \right)}{K^{2}}$. 
\end{lemma} 
The proof is provided in Appendix~\ref{proof:proof of uniform convergence}. Combining the above results will give us the result. Next we assume that the loss $\ell\left(\cdot \right) $ associates with FCN, and take it as an example in the proof. Then if the loss $\ell\left(\cdot \right) $ associates with CNN, the proof follows in the same manner.
\begin{proof}[Proof of Theorem~\ref{PD_Hessian_Classification}]
	With probability at least $1-d^{-10}$,
	\begin{align*}
	&\nabla^2 f_{n}(\bW) \notag \\
    &\succeq  \nabla^{2} f\left(\bW  \right) - \left\|\nabla^{2}f_{n}\left(\bW \right) - \nabla^{2} f(\bW) \right\| \cdot \bI \notag \\
	& \succeq \Omega\left(\frac{1}{K^{2}}\cdot \frac{\rho_{\mathrm{FCN}}\left(\sigma_K \right) }{\kappa^2 \lambda} \right) \cdot \bI - \Omega\left(C\cdot \sqrt{\frac{ d K\log n}{n} }  \right)\cdot \bI.
	\end{align*}
	As long as the sample size $n$ is set to satisfy 
	$$  C\cdot \sqrt{\frac{ dK\log n}{n} }   \leq \frac{1}{K^{2}}\cdot \frac{\rho_{\mathrm{FCN}}\left(\sigma_K \right) }{\kappa^2 \lambda} ,$$
	i.e.  $ n \geq C\cdot  dK^{5} \log^2 d \cdot \left(\frac{\kappa^{2} \lambda}{\rho_{\mathrm{FCN}}\left(\sigma_{K} \right) } \right)^{2} $, we have	
	\begin{equation*}
	\nabla^2 f_{n}(\bW)  \succeq \Omega\left(\frac{1}{K^{2}}\cdot \frac{\rho_{\mathrm{FCN}}\left(\sigma_K \right) }{\kappa^2 \lambda} \right) \cdot \bI.
	\end{equation*}
	holds for all $\bW \in \mathbb{B}\left(\bW^{\star},r_{\mathrm{FCN}} \right) $. Similarly, we have
	\begin{equation*}
	\nabla^2 f_{n}(\bW)  \preceq C \cdot \bI
	\end{equation*}
	holds for all $\bW \in \mathbb{B}\left(\bW^{\star},r_{\mathrm{FCN}} \right) $.
\end{proof}

\section{Proof of Theorem~\ref{theorem:GD_convergence_FNN}} \label{proof: performance guarantee gradient descent}
We have established that $f_{n}\left(\bW \right) $ is strongly convex in $\mathbb{B}(\bW^{\star},r)$ in Theorem \ref{PD_Hessian_Classification}. Thus there exists at most one critical point in $\mathbb{B}(\bW^{\star},r)$. The proof of Theorem \ref{theorem:GD_convergence_FNN} follows the steps below:
\begin{enumerate}
	\item We first show that the gradient $\nabla f_{n}\left(\bW \right) $ concentrates around $\nabla f\left(\bW \right)$ in $\mathbb{B}(\bW^{\star},r)$ (Lemma~\ref{lemma:gradient_convergence}), and then invoke \cite[Theorem 2]{mei2016landscape} to guarantee that there indeed exists a critical point $\widehat{\bW}_n$ in $\mathbb{B}(\bW^{\star},r)$;
	\item We next show that $\widehat{\bW}_n$ is close to $\bW^{\star}$ and gradient descent converges linearly to $\hat{\bW}_n$ with a properly chosen step size. 
\end{enumerate}

To begin, the following lemma establishes that $\nabla f_{n}\left(\bW \right) $ uniformly concentrates around $\nabla f\left(\bW \right)$.
\begin{lemma}\label{lemma:gradient_convergence}
	If the loss $\ell\left(\cdot \right) $ associates with FCN \eqref{eq:FNN:Classification_model} with $r_{\mathrm{FCN}}:=  \frac{C}{K^{\frac{1}{2}}}\cdot \frac{\rho_{\mathrm{FCN}}\left(\sigma_{K} \right) }{\kappa^{2}\lambda}$, and $\|\bw^{\star}_{k}\|_{2}\leq 1$ for all $k$, then 
	\begin{align*}
	\underset{\bW\in \mathbb{B}\left(\bW^{\star},r_{\mathrm{FCN}} \right) }{\mathrm{sup}}\ \left\|\nabla f_{n}\left(\bW \right) - \nabla f(\bW) \right\|\leq C   \sqrt{\frac{ d\sqrt{K}\log  n}{n} } 
	\end{align*}
	holds with probability at least $1-d^{-10}$, as long as $n \geq C  dK\log (dK)$. If the loss $\ell\left(\cdot \right) $ associates with CNN \eqref{eq:CNN:Classification_model}, with $r_{\mathrm{CNN}}:=C\cdot \frac{\rho_{\mathrm{CNN}}\left(\|\bw^{\star}\|_{2} \right)}{K^{2}} $ and $\|\bw^{\star}\|_{2}\leq 1$, then
	 	\begin{align}
	 \underset{\bw\in \mathbb{B}\left(\bw^{\star},r_{\mathrm{CNN}} \right) }{\mathrm{sup}}\ \|\nabla f_{n}\left(\bw \right)-\nabla f\left(\bw \right)\| \leq C\cdot \sqrt{\frac{d\log n}{n}}
	 \end{align}
	 holds with probability at least $1-d^{-10}$ as long as $n\geq C \frac{d}{K}\log\left(\frac{d}{K} \right) $. 
\end{lemma}
The proof is provided in Appendix~\ref{proof:proof of the uniform convergence of gradient}. Notice that for the population risk function $f(\bW)$, $\bW^{\star}$ is the unique critical point in $\mathbb{B}(\bW^{\star},r)$ due to local strong convexity. With Lemma~\ref{Uniform_Convergence} and Lemma~\ref{lemma:gradient_convergence}, we can invoke \cite[Theorem 2]{mei2016landscape}, which guarantees the following. 

\begin{corollary}
	If the loss $\ell\left(\cdot \right) $ associates with FCN or CNN, there exists one and only one critical point $\widehat{\bW}_n\in \mathbb{B}\left(\bW^{*},r \right)$ that satisfies $\nabla f_{n}\left(\widehat{\bW}_n \right) = \bm{0}$ correspondingly.
\end{corollary}

Again, since the proof for the case with the loss $\ell\left(\cdot \right) $ associating with FCN is the same as that for CNN, we next take FCN as an example.

We first show that $\widehat{\bW}_n$ is close to $\bW^{\star}$. By the mean value theorem, there exists $ \bW^{\prime}$ one the straight line connecting $\bW^{\star}$ and $\widehat{\bW}_n$ such that
\begin{align}
f_{n}\left(\widehat{\bW}_n \right) &= f_{n}\left(\bW^{\star} \right) + \left\langle \nabla f_{n}\left(\bW^{\star} \right), \mathrm{vec}\left( \widehat{\bW}_n -\bW^{\star} \right) \right\rangle \notag \\ 
&+ \frac{1}{2}\mathrm{vec}\left(\widehat{\bW}_n-\bW^{\star} \right)^{\top} \nabla^{2} f_{n}\left(\bW^{\prime} \right)\mathrm{vec}\left(\widehat{\bW}_n-\bW^{\star} \right) \nonumber \\
& \leq f_{n}\left(\bW^{\star} \right),   \label{bound_W_hat}
\end{align} 
where the last inequality follows from the optimality of $\widehat{\bW}_n$. By Theorem~\ref{PD_Hessian_Classification}, we have
\begin{align} 
&\frac{1}{2}\mathrm{vec}\left(\widehat{\bW}_n-\bW^{\star} \right)^{\top} \nabla^{2} f_{n}\left(\bW^{\prime} \right)\mathrm{vec}\left(\widehat{\bW}_n-\bW^{\star} \right) \notag \\ 
&\geq  \Omega\left(\frac{1}{K^{2}}\cdot \frac{\rho_{\mathrm{FCN}}\left(\sigma_K \right) }{\kappa^2 \lambda} \right) \left\|\widehat{\bW}_n - \bW^{\star}\right\|_{\mathrm{F}}^2. \label{lowerbound_eq1}
\end{align}
On the other hand, by the Cauchy-Schwarz inequality, we have 
\begin{align}
&\left|\left\langle \nabla f_{n}\left(\bW^{\star} \right), \mathrm{vec}\left( \widehat{\bW}_n -\bW^{\star} \right) \right\rangle\right| \nonumber \\
& \leq  \|\nabla f_{n}\left(\bW^{\star} \right) \|_{2} \|\widehat{\bW}_n - \bW^{\star}\| _{\mathrm{F}} \notag \\
&\leq \Omega\left(\sqrt{\frac{ d K^{1/2}\log n}{n} }  \right)\|\widehat{\bW}_n - \bW^{\star}\|_{\mathrm{F}}, \label{upperbound_eq1}
\end{align}
where the last line follows from Lemma~\ref{lemma:gradient_convergence}. Plugging \eqref{lowerbound_eq1} and \eqref{upperbound_eq1} into \eqref{bound_W_hat}, we have
\begin{equation}
\|\widehat{\bW}_n - \bW^{\star}\|_{\mathrm{F}} \leq \Omega\left(\frac{K^{\frac{9}{4}} \kappa^{2} \lambda }{\rho_{\mathrm{FCN}}\left(\sigma_{K} \right) }   \sqrt{\frac{d \log n}{n}} \right). 
\end{equation}   



Now we have established that there indeed exists a critical point in $\mathbb{B}(\bW^{\star},r_{\mathrm{FCN}})$. We can then establish the local linear convergence of gradient descent as below. Let $\bW_{t}$ be the estimate at the $t$-th iteration. Due to the update rule, we have
\begin{align}
	\bW_{t+1}-\widehat{\bW}_n &= \bW_{t} - \eta \nabla f_{n}\left(\bW_{t} \right) -  \left( \widehat{\bW}_n-\eta \nabla f_{n}\left(\widehat{\bW}_n \right) \right) \notag \\
	& = \left(\bI-\eta\int_{0}^{1}\nabla^{2}f_{n}\left( \bW\left(\gamma \right) \right)  \right) \left(\bW_{t}-\widehat{\bW}_n  \right),\notag 
\end{align}
where $\bW(\gamma) = \widehat{\bW}_n + \gamma \left(\bW_{t} - \widehat{\bW}_n \right)$ for $\gamma\in[0,1]$. If $\bW_{t}\in \mathbb{B}(\bW^{\star},r_{\mathrm{FCN}}) $, it is obvious that $\bW(\gamma) \in \mathbb{B}(\bW^{\star},r_{\mathrm{FCN}}) $, and by Theorem \ref{PD_Hessian_Classification}, we have
\begin{equation*}
H_{\min} \cdot \bI \preceq  \nabla^{2} f_n \left(\bW(\gamma)  \right)  \preceq H_{\max} \cdot \bI,
\end{equation*} 
where $H_{\min} = \Omega\left(\frac{1}{K^{2}}\cdot \frac{\rho_{\mathrm{FCN}}\left(\sigma_K \right) }{\kappa^2 \lambda} \right) $ and $H_{\max} = C$.
Therefore, we have
\begin{align}
\|\bW_{t+1}-\widehat{\bW}_n \|_{\mathrm{F}} &\leq  \|\bI-\eta\int_{0}^{1}\nabla^{2}f_{n}\left( \bW\left(\gamma \right) \right) \| \|\bW_{t}-\widehat{\bW}_n\|_{\mathrm{F}} \notag \\
&\leq \left(1-\eta H_{\min}\right) \|\bW_{t}-\widehat{\bW}_n\|_{\mathrm{F}}. 
\end{align}
Hence, by setting $\eta=\frac{1}{H_{\max}}:=\Omega\left(C \right)$, we obtain
\begin{align}
	\|\bW_{t+1}-\widehat{\bW}_n \|_{\mathrm{F}} &\leq \left(1-\frac{H_{\min}}{H_{\max}} \right) \|\bW_{t} - \widehat{\bW}_n \|_{\mathrm{F}},
\end{align}
which implies that gradient descent converges linearly to the local minimizer $\widehat{\bW}_n$.

\bibliography{bibfile_MLE}
\bibliographystyle{IEEEtran}

\newpage
\onecolumn

 \begin{center}
\large \textbf{Supplementary Materials: Additional Proofs}
\end{center}

\section{Proof of Auxiliary Lemmas} 

\subsection{Proof of Lemma~\ref{lem:smooth_pop_local}.}\label{proof: smoothness}

We prove the two claims for FCN and CNN separately as below.
\begin{itemize}
	\item \textbf{The FCN case:}	Let $\bm{\Delta} =\nabla^2 f(\bW) - \nabla^2 f(\bW^{\star})$. For each $(j,l)\in [K] \times [K]$, let $\bm{\Delta}_{j,l} \in \mathbb{R}^{d \times d}$ denote the $(j,l)$-th block of $\bm{\Delta}$. 
	Let $\ba = [\ba_1^{\top},\cdots,\ba_K^{\top}]^{\top}\in\mathbb{R}^{dK}$. By definition,
	\begin{align}\label{eq:Hessian_difference}
	\|\nabla^2 f(\bW) - \nabla^2 f(\bW^{\star})\| =   \max_{\| \ba\|=1} \ba^\top ( \nabla^2 f(\bW) - \nabla^2 f(\bW^{\star}) ) \ba  =  \max_{ \| \ba\|=1} \sum_{j=1}^K \sum_{l=1}^K  \ba_j^\top \bm{\Delta}_{j,l}  \ba_l. 
	\end{align}
	 From \eqref{eq:hessian_formula} we know that 
	\begin{align}
	\bm{\Delta}_{j,l}= \frac{\partial ^{2} f\left(\bW \right) }{\partial \bw_{j}\partial \bw_{l}} - \frac{\partial ^{2} f\left(\bW^{\star} \right) }{\partial \bw_{j}^{\star}\partial \bw_{l}^{\star}} = \mathbb{E} \left[\left( \xi_{j,l}\left(\bW \right)-\xi_{j,l}\left(\bW^{\star} \right)\right)  \cdot \bx\bx^{\top}  \right],
	\end{align}
	and then by the mean value theorem, we can further expand $\xi_{j,l}\left(\bW \right)$ as
	\begin{equation}
	\xi_{j,l}\left(\bW \right) = \xi_{j,l }\left(\bW^{\star}\right) + \sum_{k=1}^{K}\left\langle \frac{\partial \xi_{ j,l } \left(\tilde{\bW} \right) }{\partial \tilde{\bw}_{k}}, \bw_{k}-\bw_{k}^{\star} \right\rangle ,   
	\end{equation} 
	where $\tilde{\bW} = \eta\cdot \bW + \left(1-\eta \right)\bW^{\star} $ for some $\eta\in \left(0,1 \right) $. Thus we can write $\bm{\Delta}_{j,l}$ as
	\begin{align}
	\bm{\Delta}_{j,l}  =  \mathbb{E} \left[\left( \sum_{k=1}^{K}\left\langle \frac{\partial \xi_{ j,l } \left(\tilde{\bW} \right) }{\partial \tilde{\bw}_{k}}, \bw_{k}-\bw_{k}^{\star} \right\rangle \right)\cdot \bx\bx^{\top}   \right],
	\end{align}
	which can be further simplified as
	\begin{align}
	\bm{\Delta}_{j,l} =  \mathbb{E}\left[\left( \sum_{k=1}^{K} T_{j,l,k}\left\langle \bx, \bw_{k}-\bw_{k}^{\star} \right\rangle \right) \cdot \bx\bx^{\top}  \right],
	\end{align}
	by the fact that $\frac{\partial \xi_{j,l } \left(\tilde{\bW} \right) }{\partial \tilde{\bw}_{k}}$ can be written as $T_{j,l,k}\cdot \bx$, where $T_{j,l,k}\in \mathbb{R}$ is a scalar depending on $\bx$. When $j=l$, we calculate $\frac{\partial \xi_{ j,l } \left(\tilde{\bW} \right) }{\partial \tilde{\bw}_{k}}$ for illustration,
	\begin{align}\label{eq:T FCN}
		\frac{\partial \xi_{ j,j }\left(\bW \right) }{\partial \bw_{k}} = \begin{cases}
		\left(-\frac{2}{K^{2}}\frac{\phi^{\prime}\left(\bw_{j}^{\top}\bx \right)^{2} }{H\left(\bW \right)^{3} } +\frac{1}{K}\frac{\phi^{\prime\prime}\left(\bw_{j}^{\top}\bx \right) }{H\left(\bW \right)^{2} }\right) \frac{1}{K}\phi\left(\bw_{k}^{\top}\bx \right) \bx & k\neq j \\
		\left(\frac{2}{K^{2}}\left( \frac{\phi^{\prime}\left(\bw_{j}^{\top}\bx \right)\phi^{\prime\prime}\left(\bw_{j}^{\top}\bx \right)}{H\left(\bW \right)^{2}}-\frac{\phi^{\prime}\left(\bw_{j}^{\top}\bx \right)^{2} }{H\left(\bW \right)^{3} }\right)  +\frac{1}{K}\left( \frac{\phi^{\prime\prime}\left(\bw_{j}^{\top}\bx \right) }{H\left(\bW \right)^{2} } - \frac{\phi^{\prime\prime\prime}\left(\bw_{j}^{\top}\bx \right)}{H\left(\bW \right)}\right) \right) \frac{1}{K}\phi\left(\bw_{k}^{\top}\bx \right) \bx & k=j
		\end{cases},
	\end{align}
	where we have simplified the presentation by setting $y=1$, since $y$ is a binary random variable, and we will show that in either case $|T_{j,j,k}|$ is upper bounded, i.e., in this case 
\begin{align}
	|T_{j,j,k}| \leq \begin{cases}
	\max\left\lbrace \frac{2}{K^{3}}\frac{1}{H (\tilde{\bW}  )^{3}}, \frac{1}{K^{2}}\frac{1}{H (\tilde{\bW}  )^{2}} \right\rbrace & y=1 \notag \\
	\max\left\lbrace \frac{2}{K^{3}}\frac{1}{\left( 1-H (\tilde{\bW}  )\right) ^{3}}, \frac{1}{K^{2}}\frac{1}{\left( 1-H (\tilde{\bW}  )\right) ^{2}} \right\rbrace & y=0 \notag 
	\end{cases}, 
\end{align}
since $\phi\left( \cdot \right),\phi^{\prime}\left( \cdot \right),\phi^{\prime\prime}\left( \cdot \right), \phi^{\prime\prime\prime}\left( \cdot \right) $ are bounded.	
More generally, by calculating the other case we can claim that
\begin{align}\label{eq: t upper bound FCN}
	|T_{j,l,k}| \leq \max\left\lbrace \frac{2}{K^{3}}\frac{1}{H (\tilde{\bW}  )^{3}}, \frac{1}{K^{2}}\frac{1}{H (\tilde{\bW}  )^{2}}, \frac{2}{K^{3}}\frac{1}{ ( 1-H (\tilde{\bW}  ) ) ^{3}}, \frac{1}{K^{2}}\frac{1}{ ( 1-H (\tilde{\bW}  ) ) ^{2}} \right\rbrace, 
\end{align}	
holds for all $j,l,k$.		
	Then, we can upper bound $	\ba_j^\top \bm{\Delta}_{j,l}  \ba_l$ using Cauchy-Schwarz inequality, 	
	\begin{align}	
	\ba_j^\top \bm{\Delta}_{j,l}  \ba_l 
	& =  \mathbb{E}\left[\left( \sum_{k=1}^{K} T_{j,l,k}\left\langle \bx, \bw_{k}-\bw_{k}^{\star} \right\rangle \right) \cdot \left( \ba_{j}^{\top}\bx\right)  \left( \ba_{l}^{\top}\bx \right)  \right] \notag\\
	& \leq \sqrt{\mathbb{E}\left[\sum_{k=1}^{K}T_{j,l,k}^{2}\right] \cdot\mathbb{E}\left[\sum_{k=1}^{K}\left( \left\langle \bx, \bw_{k}-\bw_{k}^{\star} \right\rangle  \left( \ba_{j}^{\top}\bx\right)  \left( \ba_{l}^{\top}\bx \right)\right)^{2}  \right]}  \notag\\
	&  \leq  \sqrt{\sum_{k=1}^{K}\mathbb{E}\left[ T_{j,l,k}^{2}\right]}\cdot \sqrt{ \sum_{k=1}^{K} \|\bw_{k}-\bw_{k}^{\star}\|_{2}^{2}\cdot \|\ba_{j}\|_{2}^{2}\cdot \|\ba_{l}\|_{2}^{2}}.	 
	\end{align}
	Plug it back to \eqref{eq:Hessian_difference} we can obtain the following inequality, 
	\begin{align}
	\|\nabla^2 f(\bW) - \nabla^2 f(\bW^{\star})\| &\leq \max_{ \| \ba\|=1} \sum_{j=1}^K \sum_{l=1}^K \sqrt{\sum_{k=1}^{K}\mathbb{E}\left[ T_{j,l,k}^{2}\right]} \cdot \sqrt{ \sum_{k=1}^{K} \|\bw_{k}-\bw_{k}^{\star}\|_{2}^{2}\cdot \|\ba_{j}\|_{2}^{2}\cdot \|\ba_{l}\|_{2}^{2}}.  \label{eq:smooth_upper}
	\end{align}
	Then the problem boils down to upper bound $\mathbb{E}\left[ T_{i,j,k}^{2}\right] $, which we can apply the following lemma, whose proof can be found in Section~\ref{proof:lemma:numerator_upperbound}. 
	\begin{lemma}\label{lemma:numerator_upperbound}
		Let $\bx\sim\mathcal{N}\left(\bm{0},\bI \right)$, $t = \mathrm{max}\left\lbrace \|\bw_{1}\|_{2},\cdots \|\bw_{K}\|_{2} \right\rbrace  $ and $z\in \mathbb{Z}$ such that $z\geq 1$ , 
		for the sigmoid activation function $\phi\left(x \right) = \frac{1}{1+e^{-x}}$, the following 
		\begin{align}\label{eq:upper bound saturation term}
		&\mathbb{E}\left[\left(  \frac{1}{\frac{1}{K}\sum_{j=1}^{K}\phi\left(\bw_{j}^{\top}\bx \right)} \right)^{z} \right]  \leq C_{1}\cdot e^{ t^2 }, \quad \mathbb{E}\left[\left(  \frac{1}{\left(1- \frac{1}{K}\sum_{j=1}^{K}\phi\left(\bw_{j}^{\top}\bx \right) \right)} \right)^{z} \right]  \leq C_{2}\cdot e^{ t^2 } 
		\end{align}
		holds for some large enough constants $C_{1},C_{2}$ that depend on the constant $z$. 
	\end{lemma}
 Setting $z=4$ and $z=6$ in Lemma~\ref{lemma:numerator_upperbound}, together with \eqref{eq: t upper bound FCN} we obtain that
\begin{align}\label{eq:coefficient_upper}
\mathbb{E}\left[ T_{j,l,k}^{2}\right]  \leq \frac{C}{K^{4}} \cdot e^{\max_{1\leq i\leq k} \| \tilde{\bw}_i\|_{2}^{2}} ,
\end{align}
holds for some constant $C$. Plugging \eqref{eq:coefficient_upper} into \eqref{eq:smooth_upper}, we obtain
	\begin{align}
	&\|\nabla^2 f(\bW) - \nabla^2 f(\bW^{\star})\| \leq \frac{C}{K^{\frac{3}{2}}}e^{\|\tilde{\bW}\|_{\mathrm{F}}^{2}}\cdot \|\bW-\bW^{\star}\|_{\mathrm{F}} \cdot \max_{ \| \ba\|=1} \sum_{j=1}^K \sum_{l=1}^K \|\ba_{j}\|_{2} \|\ba_{l}\|_{2} \leq \frac{C}{K^{\frac{3}{2}}}e^{\|\tilde{\bW}\|_{\mathrm{F}}^{2}}\cdot \|\bW-\bW^{\star}\|_{\mathrm{F}}.
	\end{align}  
	Further since $e^{\max_{1\leq i\leq k} \| \tilde{\bw}_i\|_{2}^{2}}  \leq C  $ gives that $\|\bw_i-\bw_i^{\star}\|_{2} \leq 0.7$, where we have used the assumption that $\max_{1\leq i\leq k} \| \bw_i^{\star}\|_{2}^{2} \leq 1$, we conclude that  
	\begin{align}
	\|\nabla^2 f(\bW) - \nabla^2 f(\bW^{\star})\| \leq 
	& \frac{C}{K^{\frac{3}{2}}} \|\bW-\bW^{\star}\|_{\mathrm{F}}
	\end{align}	
	holds for some constant $C$.

\item \textbf{The CNN case:} according to \eqref{eq:hessian_formula_CNN}, we can calculate the upper bound of $\|\nabla^{2}f\left(\bw \right)-\nabla^{2}f\left(\bw^{\star} \right)\|$ by definition as
\begin{align}\label{eq:smoothness upperbound}
\|\nabla^{2}f\left(\bw \right)-\nabla^{2}f\left(\bw^{\star} \right)\| \leq \underset{\|\bu\|_{2}=1}{\mathrm{max}}\ \sum_{j=1}^{K}\sum_{l=1}^{K} \mathbb{E}\left[\left( g_{j,l}\left(\bw \right) - g_{j,l}\left(\bw^{\star} \right)\right) \cdot \bu^{\top}\bx^{\left( j\right) }\cdot \bx^{\left( l\right) \top}\bu\right].   
\end{align}
We then again apply the mean value theorem to $g_{j,l}\left(\bw \right)$, such that there exists $\tilde{\bw} = \eta \bw+\left(1-\eta \right)\bw $ for some $ \eta\in \left(0,1 \right) $, 
\begin{align*}
g_{j,l}\left(\bw \right) - g_{j,l}\left(\bw^{\star} \right) = \left\langle \nabla g_{j,l}\left(\tilde{\bw} \right), \bw-\bw^{\star}\right\rangle.
\end{align*} 

Similarly to the FCN case, we can write $\nabla g_{j,l}\left(\tilde{\bw} \right)$ in the form of 
\begin{align*}
\nabla g_{j,l}\left(\tilde{\bw} \right) = \sum_{k=1}^{K} S_{j,l,k}\cdot \bx^{\left( k\right) },
\end{align*}
where $S_{j,l,k}$ is a scalar that depends on $\tilde{\bw}$ and $\bx^{\left( k\right) },k=1,\cdots,K$. Again we take $j\neq l$ as an example to calculate $S_{j,l,k}$, by definition, and obtain 
\begin{align}
	K^{2}\cdot \frac{\partial g_{j,l}\left(\bw \right) }{\partial \bw} &=  \frac{\left(1-H\left(\bw \right)  \right)\phi^{\prime}\left(\bw^{\top}\bx^{\left(j \right) } \right) \phi^{\prime\prime}\left(\bw^{\top}\bx^{\left(l \right) } \right) }{\left(1-H\left(\bw \right)  \right) ^{3}}\cdot \bx^{\left(l \right)} + \frac{\left(1-H\left(\bw \right)  \right)\phi^{\prime}\left(\bw^{\top}\bx^{\left(l \right) } \right) \phi^{\prime\prime}\left(\bw^{\top}\bx^{\left(j \right) } \right) }{\left(1-H\left(\bw \right)  \right) ^{3}}\cdot \bx^{\left(j \right)} \notag \\
	&\quad - \frac{\phi^{\prime}\left(\bw^{\top}\bx^{\left(l \right) } \right) \phi^{\prime}\left(\bw^{\top}\bx^{\left(j \right) } \right) }{\left(1-H\left(\bw \right)  \right) ^{3}}\cdot \left(\frac{1}{K}\sum_{k=1}^{K} \bx^{\left(k \right)}\right), 
\end{align}
where we set $y=0$ for simplification. Then we obtain 
\begin{align}
	S_{j,l,l} = \frac{1}{K^{2}} \frac{\left(1-H\left(\bw \right)  \right)\phi^{\prime}\left(\bw^{\top}\bx^{\left(j \right) } \right) \phi^{\prime\prime}\left(\bw^{\top}\bx^{\left(l \right) } \right) }{\left(1-H\left(\bw \right)  \right) ^{3}} - \frac{1}{K^{3}} \frac{\phi^{\prime}\left(\bw^{\top}\bx^{\left(l \right) } \right) \phi^{\prime}\left(\bw^{\top}\bx^{\left(j \right) } \right) }{\left(1-H\left(\bw \right)  \right) ^{3}}. 
\end{align}
and 
\begin{align}
	|S_{j,l,l}|  \leq \frac{1}{K^{2}} \frac{1}{\left(1-H\left(\tilde{\bw} \right)  \right) ^{3}},
\end{align}
hold, where we used the fact that $0\leq H\left(\bw \right)\leq 1 $ and $\phi^{\prime}\left(\cdot \right),\phi^{\prime\prime}\left(\cdot \right)  $ are bounded. Hence in the same way, we can obtain
\begin{align}\label{eq:upperbound T cnn}
	|S_{j,l,k}| \leq \begin{cases}
	\max \left\lbrace \frac{1}{K^{2}}\frac{1}{\left(1-H\left(\tilde{\bw} \right)  \right) ^{3}},  \frac{1}{K^{2}}\frac{1}{\left(H\left(\tilde{\bw} \right)  \right) ^{3}} \right\rbrace & j\neq l \\
	\max \left\lbrace 
	\frac{1}{K}\frac{1}{ \left(1-H\left(\tilde{\bw} \right)\right)^{2}}, \frac{1}{K}\frac{1}{ \left(H\left(\tilde{\bw}\right)\right)^{2}}
	 \right\rbrace &j=l
	\end{cases}.
\end{align}

Plug these back to \eqref{eq:smoothness upperbound} we obtain
\begin{align}
\|\nabla^{2}f\left(\bw \right)-\nabla^{2}f\left(\bw^{\star} \right)\|
&\leq \underset{\|\bu\|_{2}=1}{\mathrm{max}}\ \sum_{j=1}^{K}\sum_{l=1}^{K} \mathbb{E}\left[\sum_{k=1}^{K}\left\langle S_{j,l,k}\cdot \bx^{\left( k\right) }, \bw-\bw^{\star}  \right\rangle \cdot \bu^{\top}\bx^{\left( j\right) }\cdot \bx^{\left( l\right) \top}\bu  \right] \notag\\
& = \underset{\|\bu\|_{2}=1}{\mathrm{max}}\ \sum_{j=1}^{K}\sum_{l=1}^{K} \mathbb{E}\left[\sum_{k=1}^{K} S_{j,l,k}\cdot \left( \bw-\bw^{\star} \right)^{\top}\bx^{\left( k\right) } \cdot \bu^{\top}\bx^{\left( j\right) }\cdot \bx^{\left( l\right) \top}\bu  \right] \notag\\
& \leq \underset{\|\bu\|_{2}=1}{\mathrm{max}}\ \sum_{j=1}^{K}\sum_{l=1}^{K} \sqrt{\mathbb{E}\left[\sum_{k=1}^{K} S_{j,l,k}^{2}  \right] \cdot \mathbb{E}\left[\sum_{k=1}^{K} \left(\left(\bw-\bw^{\star} \right)^{\top}\bx^{\left( k\right) }  \right)^{2}\left(\bu^{\top}\bx^{\left( j\right) } \right)^{2}\left(\bx^{\left( l\right) \top}\bu \right)^{2}  \right]} \notag \\
&\leq \underset{\|\bu\|_{2}=1}{\mathrm{max}}\ \sum_{j=1}^{K}\sum_{l=1}^{K} \sqrt{\mathbb{E}\left[\sum_{k=1}^{K} S_{j,l,k}^{2}  \right]\cdot \sum_{k=1}^{K}\|\bw-\bw^{\star}\|_{2}^{2}\cdot \|\bu\|_{2}^{2}\cdot \|\bu\|_{2}^{2}} \notag \\
& \leq   C\cdot K \cdot e^{\|\tilde{\bw}\|_{2}^{2}} \cdot \|\bw-\bw^{\star}\|_{2} , 
\end{align}
where the second inequality follows from Cauchy-Schwarz inequality, and the last inequality follows from \eqref{eq:upperbound T cnn} and Lemma~\ref{lemma:numerator_upperbound}. Further since $e^{\|\tilde{\bw}\|_{2}^{2}}  \leq C\cdot\left(1+\|\bw-\bw^{\star}\|_{2}^{2} \right) $ given that $\|\bw-\bw^{\star}\|_{2}\leq 0.7$, we conclude that
\begin{align}
\|\nabla^{2}f\left(\bw \right)-\nabla^{2}f\left(\bw^{\star} \right)\| \leq C\cdot K\cdot \|\bw-\bw^{\star}\|_{2}
\end{align}
holds for some constant $C$ and $\|\bw-\bw^{\star}\|\leq 0.7$.	 	
\end{itemize}


\subsection{Proof of Lemma~\ref{lem:Bounds_Hessian_pop_tru}}\label{proof: proof of local strong convexity}
 
We first present upper and lower bounds on the Hessian $\nabla^2 f(\bW^{\star})$ of the population risk at ground truth, and then apply Lemma~\ref{lem:smooth_pop_local} to obtain a uniform bound in the neighborhood of $\bW^{\star}$.	
	
	\begin{itemize}
			
    \item \textbf{The FCN case:} Recall
    \begin{align}
    \frac{\partial^{2} f\left(\bW^{\star} \right) }{\partial \bw_{j}^{2}}&=\mathbb{E} \left[\frac{1}{K^{2}}\cdot \left( \frac{\phi^{\prime}\left(\bw_{j}^{\star\top}\bx \right)^{2} }{ H\left(\bW^{\star} \right) \left(1-H\left(\bW^{\star}\right)  \right)   }  \right) \bx\bx^{\top} \right],\notag \\
    \frac{\partial^{2} f\left(\bW^{\star} \right) }{\partial \bw_{j} \partial \bw_{l} }&= \mathbb{E} \left[\frac{1}{K^{2}}\cdot \left( \frac{\phi^{\prime}\left(\bw_{j}^{\star\top}\bx \right) \phi^{\prime}\left(\bw_{l}^{\star\top}\bx \right)  }{ H\left(\bW^{\star}\right) \left(1- H\left(\bW^{\star}\right)\right)    }  \right) \bx\bx^{\top} \right]\notag, 
    \end{align}
    where we have applied the fact that $\mathbb{E}\left[ y|\bx\right] = H\left(\bW^{\star} \right)  $.
    Let $\ba = [\ba_1^{\top},\cdots,\ba_K^{\top}]^{\top}\in\mathbb{R}^{dK}$. Then we can write
    \begin{align}
    	\nabla^{2}f\left(\bW^{\star} \right) 
    	\succeq \left( \underset{\|\ba\|_{2} = 1}{\mathrm{min}} \ba^{\top} \nabla^{2}f\left(\bW^{\star} \right) \ba\right)  \cdot \bI = \underset{\|\ba\|_{2} = 1}{\mathrm{min}} \frac{1}{K^{2}} \mathbb{E}\left[\frac{\left(\sum_{j=1}^{K} \phi^{\prime}\left(\bw_{j}^{\star\top}\bx \right) \left( \ba_{j}^{\top}\bx\right)   \right)^{2} }{H\left(\bW^{\star} \right)  \left(1- H\left(\bW^{\star} \right)  \right)  } \right]\cdot \bI.
    \end{align}
    Since $0\leq H\left(\bW^{\star} \right)\leq 1$, we have that $H\left(\bW^{\star} \right)\left(1-H\left(\bW^{\star} \right) \right) \leq \frac{1}{4} $. Hence,
    \begin{align}
    	\nabla^{2}f\left(\bW^{\star} \right) 
    	\succeq \underset{\|\ba\|_{2} = 1}{\mathrm{min}} \frac{4}{K^{2}}
    	\mathbb{E}\left[\left(\sum_{j=1}^{K} \phi^{\prime}\left(\bw_{j}^{\star\top}\bx \right) \left( \ba_{j}^{\top}\bx\right)   \right)^{2} \right]\cdot \bI  \succeq \frac{4}{K^{2}}\cdot \frac{\rho_{\mathrm{FCN}}\left(\sigma_{K} \right) }{\kappa^{2}\lambda}\cdot \bI,\label{Lower_Hessian_ground}
    \end{align}
    where the last inequality follows from \cite[Lemmas D.4 and D.6]{zhong17a}. To derive an upper bound of $\nabla^{2}f\left(\bW^{\star}\right)$, we have
    \begin{align}\label{eq:pop_hessian_upperbound}
    \nabla^{2}f\left(\bW^{\star}\right)
    \preceq \left( \underset{\|\ba\|_{2} = 1}{\mathrm{max}} \ba^{\top} \nabla^{2}f\left(\bW^{\star} \right) \ba\right)  \cdot \bI = \underset{\|\ba\|_{2} = 1}{\mathrm{max}} \frac{1}{K^{2}} \mathbb{E}\left[\frac{\left(\sum_{j=1}^{K} \phi^{\prime}\left(\bw_{j}^{\star\top}\bx \right) \left( \ba_{j}^{\top}\bx\right)   \right)^{2} }{ \frac{1}{K^{2}}\sum_{j,l}   \phi\left(\bw_{j}^{\star\top}\bx \right) \left(1-  \phi\left(\bw_{l}^{\star\top}\bx \right)\right) } \right].
    \end{align}
    Then by Cauchy-Schwarz inequality, we have
    \begin{align}
    	\frac{\left(\sum_{j=1}^{K} \phi^{\prime}\left(\bw_{j}^{\star\top}\bx \right) \left( \ba_{j}^{\top}\bx\right)   \right)^{2} }{ \frac{1}{K^{2}}\sum_{j,l} \phi\left(\bw_{j}^{\star\top}\bx \right) \left(1-  \phi\left(\bw_{l}^{\star\top}\bx \right)\right) } \leq \frac{\left( \sum_{j=1}^{K} \phi^{\prime}\left(\bw_{j}^{\star\top}\bx \right)^{2}\right)  \cdot \left( \sum_{j=1}^{K} \left( \ba_{j}^{\top}\bx\right)^{2}  \right)   }{ \frac{1}{K^{2}}\sum_{j,l} \phi\left(\bw_{j}^{\star\top}\bx \right) \left(1-  \phi\left(\bw_{l}^{\star\top}\bx \right)\right) }. 
    \end{align}
    Further since $\phi^{\prime}\left(\bw_{j}^{\star\top}\bx \right) \leq \frac{1}{4}$, and
    \begin{align}\label{eq:lower bound H(1-H)}
    	\sum_{j,l} \phi\left(\bw_{j}^{\star\top}\bx \right) \left(1-  \phi\left(\bw_{l}^{\star\top}\bx \right)\right) \geq \sum_{j=1}^{K} \phi\left(\bw_{j}^{\star\top}\bx \right) \left(1-  \phi\left(\bw_{j}^{\star\top}\bx \right)\right)=\sum_{j=1}^{K} \phi^{\prime}\left(\bw_{j}^{\star\top}\bx \right) \geq 4 \sum_{j=1}^{K} \phi^{\prime}\left(\bw_{j}^{\star\top}\bx \right)^{2},
    \end{align}
     we obtain
    \begin{align}
    	\ba^{\top} \nabla^{2}f\left(\bW^{\star} \right) \ba \preceq \frac{1}{K^{2}} \mathbb{E}\left[\frac{CK^{2}}{4} \sum_{j=1}^{K} \left(\ba_{j}^{\top}\bx \right)^{2} \right] \label{eq:pop_hessian_upper} .
    \end{align}
    Plugging \eqref{eq:pop_hessian_upper} back to \eqref{eq:pop_hessian_upperbound}, we obtain
    \begin{align}
    \nabla^{2}f\left(\bW^{\star}\right)\preceq C\cdot \bI.
    \end{align}
    Thus together with the lower bound \eqref{Lower_Hessian_ground}, we conclude that 
    \begin{align}
    \frac{4}{K^{2}}\cdot \frac{\rho_{\mathrm{FCN}}\left(\sigma_{K} \right) }{\kappa^{2}\lambda}\cdot \bI	\preceq \nabla^{2}f\left(\bW^{\star} \right) \preceq C\cdot \bI. 
    \end{align}
    From Lemma \ref{lem:smooth_pop_local}, we have
    \begin{equation}
    \| \nabla^2 f(\bW) - \nabla^2 f(\bW^\star)\| \lesssim   \frac{C}{K^{\frac{3}{2}}} \|\bW-\bW^{\star}\|_{F}.
    \end{equation} 
    Therefore, if $\|\bW^{\star}-\bW\|_{F}\leq 0.7$ and 
    \begin{align*}
    \frac{C}{K^{\frac{3}{2}}} \cdot \|\bW-\bW^{\star}\|_{F} \leq \frac{4}{K^{2}}\cdot \frac{\rho_{\mathrm{FCN}}\left(\sigma_{K} \right) }{\kappa^{2}\lambda}, 
    \end{align*}
    i.e., if $\|\bW-\bW^{\star}\|_{F} \leq   \mathrm{min}\left\lbrace \frac{C}{K^{\frac{1}{2}}}\cdot \frac{\rho_{\mathrm{FCN}}\left(\sigma_{K} \right) }{\kappa^{2}\lambda}, 0.7\right\rbrace  $ for some constant $C$, we have
    \begin{align}
    \sigma_{\mathrm{min}}\left(\nabla^{2}f\left(\bW\right) \right) \geq \sigma_{\mathrm{min}}\left(\nabla^{2}f\left(\bW^{\star}\right) \right) - \|\nabla^{2}f\left(\bW \right) - \nabla^{2}f\left(\bW^{\star} \right)   \| &\gtrsim \frac{4}{K^{2}}\cdot \frac{\rho_{\mathrm{FCN}}\left(\sigma_{K} \right) }{\kappa^{2}\lambda} - \frac{C}{K^{\frac{3}{2}}} \|\bW-\bW^{\star}\|_{F}  \notag\\
    &\gtrsim \frac{4}{K^{2}}\cdot \frac{\rho_{\mathrm{FCN}}\left(\sigma_{K} \right) }{\kappa^{2}\lambda}.\notag     
    \end{align}
    Moreover, within the same neighborhood, by the triangle inequality we have
    \begin{align}
    \|\nabla^{2}f\left(\bW \right)\| &\leq \|\nabla^{2}f\left(\bW \right) - \nabla^{2}f\left(\bW^{\star} \right)   \| + \|\nabla^{2}f\left(\bW^{\star} \right)\| \lesssim C\notag.
    \end{align}

 \item \textbf{The CNN case:} Following from \eqref{eq:hessian_formula_CNN}, we have
    \begin{align}
    \nabla^{2} f\left(\bw^{\star} \right) &= \mathbb{E}\bigg[\frac{\frac{1}{K^{2}}\sum_{j,l}\phi^{\prime}\left(\bw^{\star\top}\bx^{\left(j \right) }  \right) \phi^{\prime}\left(\bw^{\star\top}\bx^{\left(l \right) }  \right) \bx^{\left(j \right)}  \bx^{\left(l \right)\top} }{H\left(\bw^{\star} \right) \left(1-H\left(\bw^{\star} \right) \right) } \bigg].  
    \end{align} 
    By definition, we lower bound $\nabla^{2} f\left(\bw^{\star} \right)$ by
    \begin{align}
    & \underset{\|\bu\|=1}{\mathrm{min}}\  \mathbb{E}\left[\frac{\frac{1}{K^{2}}\sum_{j,l} \phi^{\prime}\left(\bw^{\star\top}\bx^{\left(j \right) } \right) \bu^{\top}\bx^{\left(j \right) }\phi^{\prime}\left(\bw^{\star\top}\bx^{\left(l \right) } \right) \bu^{\top}\bx^{\left(l \right) } }{H\left(\bw^{\star} \right)\left(1-H\left(\bw^{\star} \right)  \right) }    \right] \cdot \bI \notag \\
     &\succeq \underset{\|\bu\|=1}{\mathrm{min}}\  \mathbb{E}\left[\frac{4}{K^{2}}  \sum_{j,l} \phi^{\prime}\left(\bw^{\star\top}\bx^{\left(j \right) } \right) \bu^{\top}\bx^{\left(j \right) }\phi^{\prime}\left(\bw^{\star\top}\bx^{\left(l \right) } \right) \bu^{\top}\bx^{\left(l \right) }  \right] \cdot \bI\notag\\
    & = \frac{4}{K^{2}}\cdot\left( \underset{\|\bu\|=1}{\mathrm{min}}\  \sum_{j\neq l} \mathbb{E}\left[\phi^{\prime}\left(\bw^{\star\top}\bx^{\left(j \right) } \right) \bu^{\top}\bx^{\left(j \right) }  \right]\cdot \mathbb{E}\left[\phi^{\prime}\left(\bw^{\star\top}\bx^{\left(l \right) } \right) \bu^{\top}\bx^{\left(l \right) }  \right]  + \sum_{j=1}^{K}\mathbb{E}\left[\left( \phi^{\prime}\left(\bw^{\star\top}\bx^{\left(j \right) } \right) \bu^{\top}\bx^{\left(j \right) } \right) ^{2} \right] \right)  \cdot \bI\notag,
    \end{align}
    
    where the last equality follows from the fact that $\bx^{\left(j \right) }$ is independent from $\bx^{\left(l \right) }$ given that $j\neq l$. Next we decompose $\bu$ as $\bu =\frac{ \bu^{\top}\bw^{\star}}{\|\bw^{\star}\|_{2}^{2}}\cdot \bw^{\star} + \left( \bu - \frac{ \bu^{\top}\bw^{\star}}{\|\bw^{\star}\|_{2}^{2}}\cdot \bw^{\star} \right) $, and calculate the expectation as
    \begin{align}
    \mathbb{E}\left[\phi^{\prime}\left(\bw^{\star\top}\bx^{\left(j \right) } \right) \bu^{\top}\bx^{\left(j \right) }  \right]
    &=\mathbb{E}\left[\phi^{\prime}\left(\bw^{\star\top}\bx^{\left(j \right) } \right)\left(  \frac{ \bu^{\top}\bw^{\star}}{\|\bw^{\star}\|_{2}^{2}}\cdot \bw^{\star} + \left( \bu - \frac{ \bu^{\top}\bw^{\star}}{\|\bw^{\star}\|_{2}^{2}}\cdot \bw^{\star} \right) \right) ^{\top}\bx^{\left(j \right) }    \right] \notag \\
    & = \mathbb{E}\left[\phi^{\prime}\left(\bw^{\star\top}\bx^{\left(j \right) } \right)\frac{ \bu^{\top}\bw^{\star}}{\|\bw^{\star}\|_{2}^{2}}\cdot \bw^{\star\top} \bx^{\left(j \right) }    \right]+ \mathbb{E}\left[\phi^{\prime}\left(\bw^{\star\top}\bx^{\left(j \right) } \right)\right]\cdot \mathbb{E} \left[ \left( \bu - \frac{ \bu^{\top}\bw^{\star}}{\|\bw^{\star}\|_{2}^{2}}\cdot \bw^{\star} \right) ^{\top}\bx^{\left(j \right) }    \right]\notag\\
    & = \frac{ \bu^{\top}\bw^{\star}}{\|\bw^{\star}\|_{2}^{2}}\mathbb{E}\left[\phi^{\prime}\left(\bw^{\star\top}\bx^{\left(j \right) } \right)\bw^{\star\top}\bx^{\left(j \right) }   \right], \notag
    \end{align}
    where the second equality follows from the independence of $\bw^{\star\top}\bx^{\left(j \right) }$ and $\left( \bu - \frac{ \bu^{\top}\bw^{\star}}{\|\bw^{\star}\|_{2}^{2}}\cdot \bw^{\star} \right) ^{\top}\bx^{\left(j \right) }$.
    Hence,
    \begin{align}
    \mathbb{E}\left[\phi^{\prime}\left(\bw^{\star\top}\bx^{\left(j \right) } \right) \bu^{\top}\bx^{\left(j \right) }  \right]\cdot \mathbb{E}\left[\phi^{\prime}\left(\bw^{\star\top}\bx^{\left(l \right) } \right) \bu^{\top}\bx^{\left(l \right) }  \right] = \left( \frac{ \bu^{\top}\bw^{\star}}{\|\bw^{\star}\|_{2}^{2}}\right) ^{2}\left( \mathbb{E}\left[\phi^{\prime}\left(z\right)z  \right]\right) ^{2} = 0,
    \end{align}
    where $z=\bw^{\star\top}\bx^{\left(j \right) } \sim\mathcal{N}\left(0,\|\bw^{\star}\|_{2}^{2} \right) $, and the last equality follows because $\phi^{\prime}\left(z\right)z = - \left( \phi^{\prime}\left(-z\right)\cdot \left( -z\right)  \right) $.	
    Similarly,
    \begin{align}
    &\mathbb{E}\left[\left( \phi^{\prime}\left(\bw^{\star\top}\bx^{\left(j \right) } \right) \bu^{\top}\bx^{\left(j \right) } \right) ^{2} \right]\notag \\
    &= \mathbb{E}\bigg[\phi^{\prime}\left(\bw^{\star\top}\bx^{\left(j \right) } \right)^{2} \cdot\left(\left( \frac{ \bu^{\top}\bw^{\star}}{\|\bw^{\star}\|_{2}^{2}}\cdot \bw^{\star\top}\bx^{\left(j \right) }\right)^{2}  + \left( \left( \bu - \frac{ \bu^{\top}\bw^{\star}}{\|\bw^{\star}\|_{2}^{2}}\cdot \bw^{\star} \right)^{\top}\bx^{\left(j \right) }\right)^{2}     \right)  \bigg] \notag \\
    & = \left(\frac{ \bu^{\top}\bw^{\star}}{\|\bw^{\star}\|_{2}^{2}} \right)^{2}\cdot \mathbb{E}\left[\phi^{\prime}\left(\bw^{\star\top}\bx^{\left(j \right) } \right)^{2} \left( \bw^{\star\top}\bx^{\left(j\right) }\right)^{2}\right]  + \left(\|\bu\|_{2}^{2} - \frac{\left(\bu^{\top}\bw^{\star} \right) ^{2}}{\|\bw^{\star}\|_{2}^{2}} \right) \cdot \mathbb{E}\left[\phi^{\prime}\left(\bw^{\star\top}\bx^{\left(j \right) } \right)^{2}  \right]. 
    \end{align}
    Together with Definition~\ref{def:rho_CNN}, we have 
    \begin{align}
    \mathbb{E}\left[\left( \phi^{\prime}\left(\bw^{\star\top}\bx^{\left(j \right) } \right) \bu^{\top}\bx^{\left(j \right) } \right) ^{2} \right]\geq \rho_{\mathrm{CNN}}\left(\|\bw^{\star}\|_{2} \right).
    \end{align}
    Hence, 
    \begin{align}
    \nabla^{2}f\left(\bw^{\star} \right) \succeq \frac{4}{K}\cdot \rho_{\mathrm{CNN}}\left(\|\bw^{\star}\|_{2} \right) \cdot \bI .  
    \end{align}
    Moreover, we apply Cauchy-Schwarz inequality and upper bound the Hessian as
    \begin{align}
    \nabla^{2}f\left(\bw^{\star} \right) &\leq \left(\underset{\|\bu\|_{2}=1}{\mathrm{max}} \bu^{\top} \nabla^{2}f\left(\bw^{\star} \right) \bu \right) \cdot \bI  \leq \underset{\|\bu\|_{2}=1}{\mathrm{max}} \mathbb{E}\left[\frac{\sum_{j=1}^{K}\left(\frac{1}{K}\phi^{\prime}\left(\bw^{\star\top}\bx^{\left(j \right) } \right)\right)^{2} \cdot \sum_{j=1}^{K} \left( \bu^{\top}\bx^{\left(j \right) }  \right)^{2} }{H\left(\bw^{\star} \right)\left(1-H\left(\bw^{\star} \right)  \right) } \right] \cdot \bI . \label{eq:Hessian_CNN_upper} 
    \end{align}
    Using \eqref{eq:lower bound H(1-H)}, i.e.,
    \begin{align}
    \frac{\frac{1}{K^{2}}\sum_{j=1}^{K}\phi^{\prime}\left(\bw^{\star \top}\bx^{\left(j \right) }\right)^{2} }{H\left(\bw^{\star} \right)\left(1-H\left(\bw^{\star} \right) \right) } \leq  \frac{1}{4},  
    \end{align}
    we upper bound the right-hand side of \eqref{eq:Hessian_CNN_upper} as
    \begin{align}
    \nabla^{2}f\left(\bw^{\star} \right) \preceq  \underset{\|\bu\|_{2}=1}{\mathrm{max}} \mathbb{E}\left[\frac{1}{4}\sum_{j=1}^{K} \left( \bu^{\top}\bx^{\left(j \right) }  \right)^{2} \right]  \cdot \bI = \frac{K}{4}\cdot \bI.
    \end{align}
    Together with the lower bound, we now conclude that
    \begin{align}
    \frac{4}{K}\cdot \rho_{\mathrm{CNN}}\left(\|\bw^{\star}\|_{2} \right) \cdot \bI \preceq \nabla^{2}f\left(\bw^{\star} \right) \preceq \frac{K}{4} \cdot \bI.
    \end{align}
    And following from \eqref{Eq:Hessian smoothness CNN} in Lemma~\ref{lem:smooth_pop_local}, we have
    \begin{align}
    \|\nabla^{2}f\left(\bw \right)-\nabla^{2}f\left(\bw^{\star} \right)\| \leq C\cdot K\cdot \|\bw-\bw^{\star}\|_{2}.
    \end{align}
    Thus if $\|\bw-\bw^{\star}\|\leq  \mathrm{min}\left\lbrace 0.7, C\cdot \frac{\rho_{\mathrm{CNN}}\left(\|\bw^{\star}\|_{2} \right)}{K^{2}} \right\rbrace $, we have
    \begin{align}
    C \cdot\frac{\rho_{\mathrm{CNN}}\left(\|\bw^{\star}\|_{2} \right)}{K}  \cdot \bI \preceq \nabla^{2}f\left(\bw \right) \preceq C \cdot K\cdot \bI.
    \end{align}   	
	\end{itemize}

\subsection{Proof of Lemma~\ref{Uniform_Convergence} }\label{proof:proof of uniform convergence}
 
We apply a covering type of argument to show that the Hessian of the empirical risk function concentrates around the Hessian of the population risk function uniformly, and the argument applies to both the loss associated with FCN and CNN. We first take the FCN case as an example and then we provide the necessary modifications for the proof of the CNN case.
\begin{itemize}
	\item \textbf{The FCN case:}					
		We adapt the analysis in \cite{mei2016landscape} to our setting. Let $N_{\epsilon}$ be the $\epsilon$-covering number of the Euclidean ball $\mathbb{B}\left(\bW^{\star},r \right)  $. Here, we omit the subscript $\mathrm{FCN}$ of $r$ for simplicity. It is known that $\log N_{\epsilon}\leq dK \log \left(3r/\epsilon \right) $ \cite{vershynin2010introduction}. Let $\mathcal{W}_{\epsilon} = \left\lbrace \bW_{1},\cdots,\bW_{N_{\epsilon}} \right\rbrace$ be the $\epsilon$-cover set with $N_{\epsilon}$ elements. For any $\bW\in \mathbb{B}\left(\bW^{\star},r \right)$, let $j\left(\bW \right) = \mathrm{argmin}_{j\in \left[N_{\epsilon} \right] }\ \|\bW-\bW_{j\left(\bW \right) }\|_{\mathrm{F}} \leq \epsilon $ for all $\bW\in \mathbb{B}\left(\bW^{\star},r \right) $.
		
		For any $\bW\in \mathbb{B} \left(\bW^{\star},r \right) $, we have 
		\begin{align}
		\left\|\nabla^{2}f_{n}\left(\bW \right) - \nabla^{2} f(\bW) \right\|
		& \leq\frac{1}{n} \left\|\sum_{i=1}^{n}\left[\nabla^{2}\ell\left(\bW;\bx_{i} \right) - \nabla^{2}\ell\left(\bW_{j\left(\bW \right) };\bx_{i} \right)  \right] \right\|  + \left\|\frac{1}{n}\sum_{i=1}^{n} \nabla^{2}\ell\left(\bW_{j\left(\bW \right) };\bx_{i} \right) - \mathbb{E}\left[\nabla^{2}\ell\left(\bW_{j\left(\bw \right) };\bx \right) \right]  \right\|\nonumber  \\
		& \quad + \left\|\mathbb{E}\left[\nabla^{2}\ell\left(\bW_{j\left(\bW \right) };\bx \right) \right]  - \mathbb{E}\left[\nabla^{2}\ell\left(\bW;\bx \right)\right] \right\|. \nonumber
		\end{align}
		Hence, we have
		\begin{align*}
		\mathbb{P}\left(\underset{\bW\in \mathbb{B} \left(\bW^{\star},r \right) }{\mathrm{sup}}\ \left\|\nabla^{2}f_{n}\left(\bW \right) - \nabla^{2} f(\bW) \right\|\geq t \right) \leq \mathbb{P}\left(A_{t} \right) + \mathbb{P}\left(B_{t} \right)+ \mathbb{P}\left(C_{t} \right),    
		\end{align*}
		where the events $A_{t}$, $B_{t}$ and $C_{t}$ are defined as
		\begin{small}
			\begin{align*}
			A_{t} &= \left\lbrace \underset{\bW\in \mathbb{B} \left(\bW^{\star},r \right)}{\mathrm{sup}}\ \frac{1}{n} \left\|\sum_{i=1}^{n}\left[\nabla^{2}\ell\left(\bW;\bx_{i} \right) - \nabla^{2}\ell\left(\bW_{j\left(\bW \right) };\bx_{i} \right)  \right] \right\| \geq \frac{t}{3}  \right\rbrace, \\
			B_{t} &= \left\lbrace \underset{\bW\in\mathcal{W}_{\epsilon} }{\mathrm{sup}}\  \left\|\frac{1}{n}\sum_{i=1}^{n} \nabla^{2}\ell\left(\bW ;\bx_{i} \right) - \mathbb{E}\left[\nabla^{2}\ell\left(\bW ;\bx \right) \right]  \right\| \geq \frac{t}{3}  \right\rbrace, \\
			C_{t} &= \left\lbrace \underset{\bW\in \mathbb{B} \left(\bW^{\star},r \right) }{\mathrm{sup}}\ \left\|\mathbb{E}\left[\nabla^{2}\ell\left(\bW_{j\left(\bW \right) };\bx \right) \right]  - \mathbb{E}\left[\nabla^{2}\ell\left(\bW ;\bx \right)\right] \right\| \geq \frac{t}{3}  \right\rbrace.    
			\end{align*} 			
		\end{small}
		
		In the sequel, we bound the terms $\mathbb{P}\left(A_{t} \right)$, $\mathbb{P}\left(B_{t} \right)$, and $\mathbb{P}\left(C_{t} \right)$, separately.
		\begin{enumerate}
			
			\item \textbf{Upper bound on $\mathbb{P}\left(B_{t} \right) $.} 
			Before continuing, we state a useful technical lemma, whose proof can be found in \cite{mei2016landscape}.
			\begin{lemma}\label{spectral_covering}
				Let $\bm{M} \in \mathbb{R}^{d\times d}$ be a symmetric $d\times d$ matrix and $V_{\epsilon}$ be an $\epsilon$-cover of unit-Euclidean-norm ball $\mathbb{B}\left(\bm{0},1 \right) $, then
				\begin{equation}
				\| \bm{M}\| \leq \frac{1}{1-2\epsilon} \underset{\bv\in V_{\epsilon}}{\sup}|\left\langle \bv, \bm{M} \bv \right\rangle |.
				\end{equation}	
			\end{lemma}
			Let $V_{\frac{1}{4}}$ be a $\left(\frac{1}{4} \right)$-cover of the ball $\mathbb{B}(\bm{0},1)= \{\bW\in\mathbb{R}^{d\times K}: \| \bW\|_{\mathrm{F}}=1\}$, where $\log|V_{\frac{1}{4}}|\leq dK\log 12$. Following from Lemma \ref{spectral_covering}, we have
			\begin{align*}
			&\left\|\frac{1}{n}\sum_{i=1}^{n} \nabla^{2}\ell\left(\bW;\bx_{i} \right) - \mathbb{E}\left[\nabla^{2}\ell\left(\bW ;\bx \right) \right]\right\| \leq 2 \underset{\bv\in V_{\frac{1}{4}}}{\sup}\left|\left\langle \bv, \left(\frac{1}{n}\sum_{i=1}^{n} \nabla^{2}\ell\left(\bW ;\bx_{i} \right) - \mathbb{E}\left[\nabla^{2}\ell\left(\bW ;\bx \right) \right] \right)\bv \right\rangle \right|.
			\end{align*}
			Taking the union bound over $\mathcal{W}_{\epsilon}$ and $V_{\frac{1}{4}}$ yields
			\begin{align}
			\mathbb{P}\left(B_{t} \right)
			\leq \mathbb{P}\left( \underset{\bW\in\mathcal{W}_{\epsilon} ,\bv\in V_{\frac{1}{4}}}{\sup} \left| \frac{1}{n}\sum_{i=1}^{n} G_{i} \right|\geq \frac{t}{6} \right)  \leq e^{dK\left( \log \frac{3r}{\epsilon} +  \log 12 \right) } \underset{\bW\in\mathcal{W}_{\epsilon} ,\bv\in V_{\frac{1}{4}}}{\mathrm{sup}} \mathbb{P} \left(\left| \frac{1}{n}\sum_{i=1}^{n} G_{i} \right|\geq \frac{t}{6}  \right), \label{sub_exponential_sub}
			\end{align} 
			where $G_i = \left\langle\bv, \left(\nabla^{2}\ell\left(\bW;\bx_{i} \right) - \mathbb{E}\left[\nabla^{2}\ell\left(\bW;\bx \right) \right] \right)\bv  \right\rangle$ and $\mathbb{E}[G_i] =0$. Let $\ba= \left[\ba_{1}^{\top},\cdots,\ba_{K}^{\top} \right]\in \mathbb{R}^{dK}$.
			Then we can show that $\|G_{i}\|_{\psi_{1}}$ is upper bounded, which we summariz as follows, and whose proof is given in Appendix~\ref{proof: lem:Hessian_sub_exponential}. 
			\begin{lemma}\label{lem:Hessian_sub_exponential}
				Suppose the loss is associated with FCN. There exists some constant $C$ such that				 
				\begin{align*}
				\|G_{i}\|_{\psi_{1}} \leq C :\equiv \tau^{2}.	
				\end{align*}
			\end{lemma}
			Applying the Bernstein inequality for sub-exponential random variables \cite[Theorem 9]{mei2016landscape} to \eqref{sub_exponential_sub}, we have that for fixed $\bW\in\mathcal{W}_{\epsilon} ,\bv\in V_{\frac{1}{4}}$,
			\begin{align}
			&\mathbb{P} \left(\left| \frac{1}{n}\sum_{i=1}^{n} \left\langle\bv, \left(\nabla^{2}\ell\left(\bW;\bx_{i} \right) - \mathbb{E}\left[\nabla^{2}\ell\left(\bW;\bx \right) \right] \right)\bv  \right\rangle \right|\geq \frac{t}{6}  \right) \leq 2\exp\left(-c\cdot n\cdot \mathrm{min}\left(\frac{t^2}{\tau^{4}},\frac{t}{\tau^{2}} \right)  \right) ,
			\end{align}	
			for some universal constant $c$. As a result, $\mathbb{P}\left(B_{t} \right)$ is upper bounded by
			\begin{equation*}
			2 \exp\left( -c\cdot n\cdot \mathrm{min}\left(\frac{t^2}{\tau^{4}},\frac{t}{\tau^{2}} \right)+ dK \log \frac{3r}{\epsilon} + dK \log 12  \right). 
			\end{equation*}
			Thus as long as
			\begin{align} 
			t&> C\cdot \mathrm{max} \bigg \{  \sqrt{\frac{\tau^{4}\left(dK \log \frac{36r}{\epsilon}+\log \frac{4}{\delta} \right) }{n}},
			\frac{\tau^{2}\left(dK\log \frac{36r}{\epsilon} + \log\frac{4}{\delta} \right) }{n}  \bigg \}    \label{requirement_n_C} 
			\end{align}
			for some large enough constant $C$, we have $\mathbb{P}\left(B_{t} \right)\leq \frac{\delta}{2} $.

			\item \textbf{Upper bound on $\mathbb{P}\left(A_{t} \right) $ and $\mathbb{P}\left(C_{t} \right) $.} These two events will be bounded in a similar way. We first present the following useful Lemma, whose proof is provided in Appendix~\ref{proof: proof of j_star}
			
			\begin{lemma} \label{J_star_bound}
				Suppose the loss is associated with FCN. There exists some constant $C$ such that
				\begin{align}
				\mathbb{E} \left[\underset{\bW\neq \bW^{\prime} \in \mathbb{B}\left(\bW^{\star},r\right) }{\mathrm{sup}} \ \frac{\|\nabla^{2}\ell\left(\bW,\bx \right)- \nabla^{2}\ell\left(\bW^{\prime},\bx \right)\| }{\|\bW-\bW^{\prime}\|_{\mathrm{F}}} \right]  \leq C\cdot d\sqrt{K}.
				\end{align}
			\end{lemma}
			
			Consider the event $C_{t}$ first. We derive
			\begin{align}
			&\underset{\bW\in \mathbb{B} \left(\bW^{\star},r \right) }{\mathrm{sup}}\ \|\mathbb{E}\left[\nabla^{2}\ell\left(\bW_{j\left(\bW \right) };\bx \right) \right]  - \mathbb{E}\left[\nabla^{2}\ell\left(\bW ;\bx \right)\right] \|\notag \\
			&\leq \underset{\bW\in \mathbb{B} \left(\bW^{\star},r \right) }{\mathrm{sup}}\ \frac{\|\mathbb{E}\left[\nabla^{2}\ell\left(\bW_{j\left(\bW \right) };\bx \right) \right]  - \mathbb{E}\left[\nabla^{2}\ell\left(\bW ;\bx \right)\right] \|}{\| \bW-\bW_{j\left( \bW\right) }\|_{\mathrm{F}}}\cdot \underset{\bW\in \mathbb{B}\left(\bW^{\star},r \right) }{\sup} \| \bW-\bW_{j\left( \bW\right) }\|_{\mathrm{F}} \nonumber \\	
			&  \leq C\cdot d\sqrt{K} \cdot \epsilon.   
			\end{align}
			Therefore, $C_t$ holds as long as
			\begin{equation} \label{requirement_t_B}
			t \geq C\cdot d\sqrt{K} \cdot \epsilon   .
			\end{equation}
			
			We can bound the event $A_{t}$ as below.
			\begin{small}
				\begin{align}
				& \mathbb{P}\left(  \underset{\bW\in \mathbb{B} \left(\bW^{\star},r \right)}{\mathrm{sup}}\ \frac{1}{n} \left\|\sum_{i=1}^{n}\left[\nabla^{2}\ell\left(\bW;\bx_{i} \right) - \nabla^{2}\ell\left(\bW_{j\left(\bW \right) };\bx_{i} \right)  \right] \right\| \geq \frac{t}{3} \right) \nonumber \\
				&\leq \frac{3}{t} \mathbb{E}\left[\underset{\bW\in \mathbb{B} \left(\bW^{\star},r \right) }{\mathrm{sup}}\ \left\|\frac{1}{n}\sum_{i=1}^{n}\left[\nabla^{2}\ell\left(\bW;\bx_{i} \right) - \nabla^{2}\ell\left(\bW_{j\left(\bW \right) };\bx_{i} \right)  \right]  \right\|  \right]   \label{markov} \\
				& \leq \frac{3}{t} \mathbb{E}\left[\underset{\bW\in \mathbb{B} \left(\bW^{\star},r \right) }{\mathrm{sup}}\ \left\|  \nabla^{2}\ell\left(\bW;\bx_{i} \right) - \nabla^{2}\ell\left(\bW_{j\left(\bW \right) };\bx_{i} \right)   \right\|  \right] \nonumber  \\
				&  \leq \frac{3}{t} \mathbb{E}\left[ \underset{\bW \in \mathbb{B}\left(\bW^{\star},r \right) }{\sup}\ \frac{\|\nabla^{2}\ell\left(\bW;\bx_{i} \right) - \nabla^{2}\ell\left(\bW_{j\left(\bW \right) };\bx_{i} \right) \|}{\|\bW-\bW_{j\left(\bW \right) }\|_{\mathrm{F}}}    \right]\cdot\underset{\bW\in \mathbb{B}\left(\bW^{\star},r \right) }{\sup}\ \|\bW-\bW_{j\left(\bW \right) }\|_{\mathrm{F}} \nonumber \\
				& \leq \frac{C\cdot d\sqrt{K}\cdot\epsilon}{t}   
				\end{align}
			\end{small}
			where \eqref{markov} follows from the Markov inequality. Thus, taking 
			\begin{equation}\label{requirement_t_A}
			t\geq \frac{6\epsilon\cdot C\cdot d\sqrt{K}}{\delta}
			\end{equation} 
			ensures that $\mathbb{P}\left(A_{t} \right)\leq \frac{\delta}{2} $.
						
			\item \textbf{Final step.} Let $\epsilon = \frac{ \delta \tau^{2}}{ C\cdot d\sqrt{K} \cdot n dK}$ and $\delta=d^{-10}$.
			Plugging $\epsilon$ and $\delta$ into \eqref{requirement_n_C} we need
			\begin{align*}
			t&> \tau^{2}\cdot \mathrm{max}\bigg \{ \frac{1}{ndK},C \cdot  \sqrt{\frac{\left(dK \log (36r n d^{11}K)+\log \frac{4}{\delta} \right) }{n}}, \frac{\left(dK\log (36r n d^{11}K) + \log\frac{4}{\delta} \right) }{n} \bigg \}.
			\end{align*}
			The middle term can be bounded as
			\begin{align}
			\frac{dK \log (36r n d^{11}K)+10 \log d}{n} \leq \frac{dK\log n}{n} + \frac{dK\log 36r}{n} + \frac{11dK  \log dK}{n} + \frac{10\log d}{n}.\notag
			\end{align}	
			If $n\geq C \cdot dK\log\left(  dK\right) $ for some large enough constant $C$, the first term $dK\log n$ dominants and is on the order of $dK \log \left( dK\right) $. Moreover, it decreases as $n$ increases when $n\geq 3$. Thus we can set
			\begin{equation}
			t \geq \tau^{2} \sqrt{\frac{\left(dK \log (36r n d^{11}K)+\log \frac{4}{\delta} \right) }{n}}
			\end{equation}
			which holds as $t\geq C' \cdot \tau^{2} \sqrt{\frac{dK \log n}{n}}$ for some constant $C'$. By setting $t:= C \tau^2 \sqrt{\frac{dK \log n}{n}}$ for sufficiently large $C$, as long as $n\geq C' \cdot dK\log dK$,
			\begin{align}
			\mathbb{P} \left(\underset{\bW\in \mathbb{B} \left(\bW^{\star},r \right) }{\mathrm{sup}}\ \|\nabla^{2}f_{n}\left(\bW \right) - \nabla^{2} f(\bW) \|\geq C \tau^2 \sqrt{\frac{dK \log n}{n}} \right) 
			&\leq d^{-10}.
			\end{align}		
		\end{enumerate}

\item \textbf{The CNN case:}
          If the loss is associated with CNN, we redefine $G_{i}$ as $G_i = \left\langle\bv, \left(\nabla^{2}\ell\left(\bw;\bx_{i} \right) - \mathbb{E}\left[\nabla^{2}\ell\left(\bw;\bx \right) \right] \right)\bv  \right\rangle$ and we show the following Lemmas whose proof is given in Appendix~\ref{proof:lem:Hessian_subexponential_CNN} and Appendix~\ref{proof:J_star_bound_CNN} .  		
		\begin{lemma}\label{lem:Hessian_subexponential_CNN}
			Suppose the loss is associated CNN. There exists some constant $C$ such that 
			\begin{align}
			\|G_{i}\|_{\psi_{1}} \leq C\cdot K^{2} :\equiv \tau^{2}.
			\end{align}
		\end{lemma}
	\begin{lemma} \label{J_star_bound_CNN}
		Suppose the loss is associated with CNN. There exists some constant $C$ such that
		\begin{align}
		\mathbb{E} \left[\underset{\bW\neq \bW^{\prime} \in \mathbb{B}\left(\bW^{\star},r\right) }{\mathrm{sup}} \ \frac{\|\nabla^{2}\ell\left(\bW,\bx \right)- \nabla^{2}\ell\left(\bW^{\prime},\bx \right)\| }{\|\bW-\bW^{\prime}\|_{\mathrm{F}}} \right]  \leq C\cdot d\sqrt{K}.
		\end{align}
	\end{lemma}
		Following argument similar to the proof of Lemma~\ref{Uniform_Convergence}, we can obtain the following concentration inequality:
		\begin{align}
		\underset{\bw\in \mathbb{B}\left(\bw^{\star},r \right) }{\mathrm{sup}} \|\nabla^{2}f_{n}\left(\bw \right)-\nabla^{2}f\left(\bw \right)\| \leq C\cdot K^{2}\sqrt{\frac{\frac{d}{K}\cdot\log n}{n}},
		\end{align}
		holds with probability at least $1-d^{-10}$, as long as the sample complexity $n\geq C\cdot \frac{d}{K}\log\left(\frac{d}{K} \right) $.	
	
\end{itemize}

\subsection{Proof of Lemma~\ref{lemma:gradient_convergence}}\label{proof:proof of the uniform convergence of gradient}
 
In order to proceed we need the following Lemma~\ref{Gradient_statistical} whose proof is given in Appendix~\ref{proof:proof of lemma Gradient_statistical}.  
\begin{lemma}\label{Gradient_statistical}
	Suppose the loss is associated with FCN. Let $\bu$ be a fixed unit norm vector $\bu = \left[\bu_{1}^{\top},\cdots,\bu_{K}^{\top} \right]\in \mathbb{R}^{dK}$ with $\|\bu\|_{2}=1$. Then we have
	\begin{equation*} 
	\| \bu^{\top} \nabla \ell\left(\bW;\bx \right) \|_{\psi_{2}} \leq  \sqrt{K}.
	\end{equation*}
	Suppose the loss is associated with CNN. Let $\bu$ be a fixed unit norm vector $\bu\in \mathbb{R}^{m}$ with $\|\bu\|_{2}=1$. Then \begin{align*}
	\|\left\langle\bu,\nabla \ell\left(\bw \right)  \right\rangle \|_{\psi_{2}} \leq C\cdot K.
	\end{align*}
\end{lemma}	

Following argument (details omitted) similar to the proof of Lemma \ref{Uniform_Convergence}, and applies Lemma \ref{Gradient_statistical}, for the loss associated with FCN, we can get the following concentration inequality
\begin{equation}\label{Gradient_Concentration}
\sup_{\bW\in \mathbb{B}\left(\bW^{\star},r_{\mathrm{FCN}} \right)}  \|\nabla f_{n}\left(\bW \right) - \nabla f\left(\bW \right)  \|_{2} \leq C\cdot  \sqrt{\frac{ d\sqrt{K} \log n}{n} } 
\end{equation}
with probability at least $1-d^{-10}$, as long as the sample size $n\geq C\cdot dK \log (dK)$. For the loss associated with CNN, we obtain	
\begin{align}
\underset{\bw\in \mathbb{B}\left(\bw^{\star},r_{\mathrm{CNN}} \right) }{\mathrm{sup}} \|\nabla f_{n}\left(\bw \right)-\nabla f\left(\bw \right)\| \leq C\cdot\sqrt{K} \sqrt{\frac{\frac{d}{K}\log n}{n}} = C\cdot \sqrt{\frac{d\log n}{n}},
\end{align}
with probability at least $1-d^{-10}$ as long as $n\geq C\cdot \frac{d}{K}\log\left(\frac{d}{K} \right) $.
 
\subsection{Proof of Lemma~\ref{lemma:numerator_upperbound}} \label{proof:lemma:numerator_upperbound}

	We take the first term in \eqref{eq:upper bound saturation term} as an example, since the second term follows exactly in the same way. We first derive
	\begin{align}
	\mathbb{E}\left[\left( \frac{1}{K}\sum_{i=1}^{K} \phi\left(\bw_{i}^{\top}\bx \right) \right)^{-z}  \right] \leq \mathbb{E}\left[ \frac{1}{K}\sum_{i=1}^{K} \left(\phi\left(\bw_{i}^{\top}\bx \right)   \right)^{-z}   \right] ,
	\end{align}
	which follows from the fact that $f\left(x \right) = x^{-z} $ is convex for $x> 0$ and $z\geq 1$. 
	Further since $\frac{1}{\phi\left(x \right) } = 1+e^{-x} $, and $g = \bw_{i}^{\top}\bx \sim \mathcal{N}\left(0,\sigma_{i}^{2}=\|\bw_{i}\|^{2}_{2} \right)$, we can exactly calculate the summands in the above equation as follows:
	\begin{align*}
	\mathbb{E}\left[\phi\left(g \right)^{-z}\right] 
	= \mathbb{E}\left[ \sum_{l=0}^{z} \binom{z}{l} e^{-lg} \right] 
	= \sum_{l=0}^{z} \binom{z}{l}  e^{\left( \frac{\sigma_{i}^2 l^{2}}{2}\right) }, 
	\end{align*}
	where we use the fact that $g$ is a Gaussian random variable. Hence, we conclude that
	for $t = \mathrm{max}\left(\|\bw_{1}\|_{2},\cdots,\|\bw_{K}\|_{2} \right) $ and $p\geq 1$,  
	\begin{align}
	\mathbb{E}\left[\left(  \frac{1}{\frac{1}{K}\sum_{i=1}^{K}\phi\left(\bw_{i}^{\top}\bx \right)   } \right)^{z} \right]  \leq C\cdot e^{t^{2}},
	\end{align}
	holds for some constant $C$ depending on $z$.

\subsection{Proof of Lemma~\ref{lem:Hessian_sub_exponential}}\label{proof: lem:Hessian_sub_exponential}
 The sub-exponential norm of $G_i$ can be bounded as 
\begin{align*}
\|G_{i}\|_{\psi_{1}} \leq \|\left\langle \bu, \nabla^{2} \ell \left(\bW;z \right)  \bu \right\rangle \|_{\psi_{1}} + \|\nabla^{2} f \left(\bW;z \right)\|,	
\end{align*}
where $\|\nabla^{2} f \left(\bW;z \right)\|$ is upper bounded by $C$ due to Lemma \ref{lem:Bounds_Hessian_pop_tru}. Denote the $\left(j,l \right)$-th block of $\nabla^{2} \ell \left(\bW;z \right)$ as $\xi_{j,l}\cdot \bx\bx^{\top}$. We can derive
\begin{align}\label{Hessian_Psi1}
\|\left\langle \bu, \nabla^{2} \ell \left(\bW;z \right)  \bu \right\rangle \|_{\psi_{1}} \leq  \sum_{j=1}^{K}\sum_{l=1}^{K} \|\xi_{j,l}\cdot \bu_{j}^{\top}\bx\bx^{\top}\bu_{l} \|_{\psi_{1}}  \leq  \sum_{j=1}^{K}\sum_{l=1}^{K} \underset{t\geq 1}{\mathrm{sup}}\quad  t^{-1}\left(\mathbb{E}\left|\xi_{j,l}\cdot \bu_{j}^{\top}\bx\bx^{\top}\bu_{l}  \right|^{t} \right)^{\frac{1}{t}}.  
\end{align}	
Next we show that $\xi_{j,l}$ is upper bounded by some constant for all $j$ and $l$.
\begin{itemize}
	\item For $j\neq l$, 	
	\begin{align}
	|\xi_{j,l}| &= \left| \frac{1}{K^{2}} \frac{\phi^{\prime} \left(\bw_{j}^{\top}\bx \right)\phi^{\prime} \left(\bw_{l}^{\top}\bx \right)\cdot\left(  H\left(\bW \right)^{2} +y - 2y\cdot H\left(\bW \right) \right)  }{\left(H\left(\bW\right)\left(1-H\left(\bW\right) \right)  \right)^{2} }\right| = \begin{cases}
	\frac{1}{K^{2}} \frac{\phi^{\prime} \left(\bw_{j}^{\top}\bx \right)\phi^{\prime} \left(\bw_{l}^{\top}\bx \right)}{\left(1-H\left(\bW\right) \right)^{2}} & y=0 \\
	\frac{1}{K^{2}} \frac{\phi^{\prime} \left(\bw_{j}^{\top}\bx \right)\phi^{\prime} \left(\bw_{l}^{\top}\bx \right)}{H\left(\bW\right)^{2}} & y=1
	\end{cases}.
	\end{align}
	Moreover,
	\begin{align}
		\frac{1}{K^{2}} \frac{\phi^{\prime} \left(\bw_{j}^{\top}\bx \right)\phi^{\prime} \left(\bw_{l}^{\top}\bx \right)}{\left(1-H\left(\bW\right) \right)^{2}} \leq \frac{\phi^{\prime} \left(\bw_{j}^{\top}\bx \right)\phi^{\prime} \left(\bw_{l}^{\top}\bx \right)}{\left( 1-\phi\left(\bw_{j}^{\top}\bx \right)\right) \left( 1- \phi\left(\bw_{l}^{\top}\bx \right)\right)} \leq \phi \left(\bw_{j}^{\top}\bx \right)\phi \left(\bw_{l}^{\top}\bx \right) \leq 1,
	\end{align}
	where the first inequality holds due to the following fact, 
	\begin{align}
\left( 1-H\left(\bW \right)\right)^{2}  =\left( 1- \frac{1}{K}\sum_{j=1}^{K}\phi\left(\bw_{j}^{\top}\bx \right)\right)^{2}  \geq \frac{1}{K^{2}} \left( 1-\phi\left(\bw_{j}^{\top}\bx \right)\right) \left( 1- \phi\left(\bw_{l}^{\top}\bx \right)\right) \notag ,  
	\end{align}
	the second inequality follows because $\phi\left(x \right)\left(1-\phi\left(x \right) \right) =  \phi^{\prime}\left(x \right)$. Similarly, we can show that 
	\begin{align}
		\frac{1}{K^{2}} \frac{\phi^{\prime} \left(\bw_{j}^{\top}\bx \right)\phi^{\prime} \left(\bw_{l}^{\top}\bx \right)}{H\left(\bW\right)^{2}} \leq 1.
	\end{align}
	 Thus for $j\neq l$, $|\xi_{j,l}| \leq 1$ holds.	
	\item For $j=l$,
	\begin{align}
	|\xi_{j,j}| \leq \left|\frac{1}{K^{2}} \frac{\phi^{\prime} \left(\bw_{j}^{\top}\bx \right)\phi^{\prime} \left(\bw_{j}^{\top}\bx \right)\cdot\left(  H\left(\bW \right)^{2} +y - 2y\cdot H\left(\bW \right) \right)  }{\left(H\left(\bW\right)\left(1-H\left(\bW\right) \right)  \right)^{2} }\right| +\left|\frac{1}{K} \frac{\phi^{\prime\prime}\left(\bw_{j}^{\top}\bx \right)\left(y - H\left(\bW \right)  \right)  }{H\left(\bW\right)\left(1-H\left(\bW\right) \right)}\right|.
	\end{align}
	For the second term in the above equation, we have
	\begin{align*}
	\left|\frac{1}{K}\frac{\phi^{\prime\prime}\left(\bw_{j}^{\top}\bx \right)\left(y - H\left(\bW \right)  \right)  }{H\left(\bW\right)\left(1-H\left(\bW\right) \right)}\right| = \begin{cases}
	\frac{1}{K}\frac{\phi^{\prime\prime}\left(\bw_{j}^{\top}\bx \right) }{\left(1-H\left(\bW\right) \right)}\leq  1 & y=0 \\
	\frac{1}{K}\frac{\phi^{\prime\prime}\left(\bw_{j}^{\top}\bx \right) }{H\left(\bW\right)}\leq 1 &y=1
	\end{cases},
	\end{align*}
	which follows from the fact that the second derivative is $\phi^{\prime\prime}\left(x \right) =  \phi\left(x \right)\left(1-\phi\left(x \right) \right)\left(1-2\phi\left(x \right) \right) $, the absolute value of which can be upper bounded by $\phi\left(x \right)$ or $1-\phi\left(x \right)$. 
\end{itemize}
 	
Hence,
\begin{align}
\left \|\left\langle \bu, \nabla^{2} \ell \left(\bW;z \right)  \bu \right\rangle \right\|_{\psi_{1}} &\leq C \cdot  \sum_{j=1}^{K}\sum_{l=1}^{K} \underset{t\geq 1}{\mathrm{sup}}\quad  t^{-1}  \left( \sqrt{\mathbb{E}\left[\left(\bu_{j}^{\top}\bx \right)^{2t}\right] }\cdot \sqrt{\mathbb{E}\left[\left(\bu_{l}^{\top}\bx \right)^{2t}\right] }\right)^{\frac{1}{t}}   \notag\\	
& \leq C\cdot  \sum_{j=1}^{K}\sum_{l=1}^{K} \|\bu_{j}\|_{2} \|\bu_{l}\|_{2}\cdot  \underset{t\geq 1}{\mathrm{sup}}\quad  t^{-1} \left(\left(2t-1 \right)!!  \right)^{\frac{1}{t}} \notag \\
&\leq C   :\equiv \tau^{2} 	,
\end{align}
where the last inequality holds because 
\begin{align}
	&\underset{t\geq 1}{\mathrm{sup}}\quad  t^{-1} \left(\left(2t-1 \right)!!  \right)^{\frac{1}{t}}\leq \underset{t\geq 1}{\mathrm{sup}}\quad  t^{-1}\left((2t)^{t} \right)^{\frac{1}{t}} \leq 2, \notag\\
	&\sum_{j=1}^{K}\sum_{l=1}^{K} \|\bu_{j}\|_{2} \|\bu_{l}\|_{2}\leq \sum_{j=1}^{K}\sum_{l=1}^{K} \frac{ \|\bu_{j}\|_{2}^{2} +  \|\bu_{l}\|_{2}^{2}}{2}=\frac{1}{2}.
\end{align}
Thus, we conclude
\begin{align*}
\|G_{i}\|_{\psi_{1}} \leq C   :\equiv \tau^{2}. 	
\end{align*}	



\subsection{Proof of Lemma~\ref{lem:Hessian_subexponential_CNN}}\label{proof:lem:Hessian_subexponential_CNN}
Again the sub-exponential norm of $G_i$ can be bounded as 
\begin{align*}
\|G_{i}\|_{\psi_{1}} \leq \|\left\langle \bu, \nabla^{2} \ell \left(\bw;z \right)  \bu \right\rangle \|_{\psi_{1}} + \|\nabla^{2} f \left(\bw;z \right)\|,	
\end{align*}
where $\|\nabla^{2} f \left(\bW;z \right)\|$ is upper bounded by $C\cdot K$ due to Lemma \ref{lem:Bounds_Hessian_pop_tru}.
Applying the triangle inequality, the sub-exponential norm of $\left\langle \bu,\nabla^{2}\ell\left(\bw \right)\bu  \right\rangle $ can be bounded as
\begin{align}\label{eq:sub-exponential}
\|\left\langle \bu,\nabla^{2}\ell\left(\bw \right)\bu  \right\rangle\|_{\psi_{1}} \leq \sum_{j\neq l} \|g_{j,l}\left(\bw \right) \bu^{\top}\bx^{\left( j\right) } \bu^{\top}\bx^{\left( l\right) } \|_{\psi_{1}}+ \sum_{j= l} \|g_{j,l}\left(\bw \right) \bu^{\top}\bx^{\left( j\right) } \bu^{\top}\bx^{\left( l\right) } \|_{\psi_{1}}.
\end{align}
Hence, we have
\begin{align*}
\left |\frac{1}{K^{2}}\frac{H\left(\bw \right)^{2}+y-2y\cdot H\left(\bw \right)}{\left(H\left(\bw \right)\left(1-H\left(\bw \right) \right)  \right)^{2} } \phi^{\prime}\left(\bw^{\top}\bx^{\left(j \right) } \right)\phi^{\prime}\left(\bw^{\top}\bx^{\left(l \right) } \right) \right|= \begin{cases}
\frac{1}{K^{2}}\frac{\phi^{\prime}\left(\bw^{\top}\bx^{\left(j \right) } \right)\phi^{\prime}\left(\bw^{\top}\bx^{\left(l \right) } \right)}{H\left(\bw \right)^2} \leq 1 & y=1 \\
\frac{1}{K^{2}}\frac{\phi^{\prime}\left(\bw^{\top}\bx^{\left(j \right) } \right)\phi^{\prime}\left(\bw^{\top}\bx^{\left(l \right) } \right)}{(1-H\left(\bw \right))^{2}} \leq 1 & y=0 	
\end{cases},
\end{align*}

\begin{align*}
\left|\frac{1}{K}\frac{y-H\left(\bw \right)}{H\left(\bw \right)\left(1-H\left(\bw \right) \right) } \phi^{\prime}\left(\bw^{\top}\bx^{\left(j \right) } \right) \right| = \begin{cases}
\frac{1}{K}\frac{\phi^{\prime}\left(\bw^{\top}\bx^{\left(j \right) } \right)}{H\left(\bw \right)} \leq 1 & y=1 \\
\frac{1}{K}\frac{\phi^{\prime}\left(\bw^{\top}\bx^{\left(j \right) } \right)}{1-H\left(\bw \right)} \leq 1 & y=0
\end{cases}.
\end{align*}
Plugging it back to \eqref{eq:sub-exponential}, we obtain
\begin{align}
\|\left\langle \bu,\nabla^{2}\ell\left(\bw \right)\bu  \right\rangle\|_{\psi_{1}} \leq \sum_{j\neq l}\|\left(\bu^{\top}\bx^{\left(j \right) } \right) \left(\bu^{\top}\bx^{\left(l \right) }\right) \|_{\psi_{1}} + \sum_{j=1}^{K}\|\left(\bu^{\top}\bx^{\left(j \right) } \right)^{2}\|_{\psi_{1}}\leq C\cdot K^{2}.
\end{align}


\subsection{Proof of Lemma~\ref{J_star_bound} }\label{proof: proof of j_star}


	As noted before, we can write the $\left(j,l \right) $-th block of $\nabla^{2} \ell \left(\bW;\bz \right)$ as $\xi_{j,l}\left(\bW \right) \bx\bx^{\top}$.	
	Then we can obtain the following bound,
	\begin{align}
	\|\nabla^{2} \ell \left(\bW;z \right) - \nabla^{2} \ell \left(\bW^{\prime};z \right)\| \leq \sum_{j=1}^{K}\sum_{l=1}^{K} |\xi_{j,l}\left(\bW \right)-\xi_{j,l}\left(\bW^{\prime} \right) |\cdot \|\bx\bx^{\top}\|.
	\end{align}
	Using the same method as shown in the proof of Lemma~\ref{lem:smooth_pop_local}, we can upper bound $|\xi_{j,l}\left(\bW \right)-\xi_{j,l}\left(\bW^{\prime} \right) |$ as
	\begin{align}
	|\xi_{j,l}\left(\bW \right)-\xi_{j,l}\left(\bW^{\prime} \right) | \leq \left(\max_{k} |T_{j,l,k}| \right)  \cdot \|\bx\|_{2}\cdot \sqrt{K} \cdot \|\bW-\bW^{\prime}\|_{F}, \notag
	\end{align}
	where following from \eqref{eq: t upper bound FCN}, 
	\begin{align}
	|T_{j,l,k}| \leq \max\left\lbrace \frac{2}{K^{3}}\frac{1}{H\left(\bW \right)^{3}}, \frac{1}{K^{2}}\frac{1}{H\left(\bW \right)^{2}}, \frac{2}{K^{3}}\frac{1}{\left( 1-H\left(\bW \right)\right) ^{3}}, \frac{1}{K^{2}}\frac{1}{\left( 1-H\left(\bW \right)\right) ^{2}} \right\rbrace. 
	\end{align}		 
	And thus, if $\|\bW-\bW^{\prime}\|_{F} \leq 0.7$ we have
	\begin{align}
	\mathbb{E} \left[ \underset{\bW\neq \bW^{\prime}}{\mathrm{sup}} \frac{\|\nabla^{2} \ell \left(\bW \right) - \nabla^{2} \ell \left(\bW^{\prime}\right) \|}{\|\bW-\bW^{\prime}\|_{F}} \right] \leq \sqrt{K} \cdot K^{2}\cdot \mathbb{E} \left[ \left(\max_{j,l,k} |T_{j,l,k}| \right)\cdot \|\bx\|_{2} \cdot \|\bx\bx^{\top}\|  \right] \leq C\cdot d \sqrt{K} .
	\end{align}
	Thus we only need to set $J^{\star}\geq C\cdot d \sqrt{K}  $ for some large enough $C$.

\subsection{Proof of Lemma~\ref{J_star_bound_CNN}}\label{proof:J_star_bound_CNN}
Following from \eqref{eq:hessian_formula_CNN} we can write
\begin{align}
\|\nabla^{2}\ell\left(\bw \right)-\nabla^{2}\ell\left(\bw^{\prime} \right)\| \leq  \sum_{j=1}^{K}\sum_{l=1}^{K} | g_{j,l}\left(\bw \right) - g_{j,l}\left(\bw^{\prime} \right)| \cdot \|\bx^{\left( j\right) } \bx^{\left( l\right) \top}\|.   
\end{align}
Similarly, the analysis in the proof of Lemma~\ref{lem:smooth_pop_local} implies that
\begin{align}
| g_{j,l}\left(\bw \right) - g_{j,l}\left(\bw^{\prime} \right)| \leq \left(\max_{k} |S_{j,l,k}| \right) \cdot \sqrt{K}\|\bx\|_{2}\cdot \|\bw-\bw^{\prime}\|_{2},
\end{align}
where we upper-bound $S_{j,l,k}$ in \eqref{eq:upperbound T cnn} as
\begin{align}
|S_{j,l,k}| \leq \begin{cases}
\max \left\lbrace \frac{1}{K^{2}}\frac{1}{\left(1-H\left(\bw \right)  \right) ^{3}},  \frac{1}{K^{2}}\frac{1}{\left(H\left(\bw \right)  \right) ^{3}} \right\rbrace & j\neq l \\
\max \left\lbrace 
\frac{1}{K}\frac{1}{ \left(1-H\left(\bw \right)\right)^{2}}, \frac{1}{K}\frac{1}{ \left(H\left(\bw \right)\right)^{2}}
\right\rbrace &j=l
\end{cases}.
\end{align}
Hence, if $\|\bw-\bw^{\prime}\|_{2}\leq 0.7$, we have
\begin{align}
\mathbb{E} \left[ \underset{\bw\neq \bw^{\prime}}{\mathrm{sup}} \frac{\|\nabla^{2} \ell \left(\bw \right) - \nabla^{2} \ell \left(\bw^{\prime}\right) \|}{\|\bw-\bw^{\prime}\|_{F}} \right] \leq \sqrt{K}\cdot \sum_{j=1}^{K}\sum_{l=1}^{K} \mathbb{E}\left[\left(\max_{k} |S_{j,l,k}| \right) \cdot \|\bx\|_{2}\cdot \|\bx^{\left( j\right) } \bx^{\left( l\right) \top}\|  \right]\leq C\cdot d\sqrt{K}. 
\end{align}
Thus, in this case we can set $J^{\star}\geq C\cdot d\sqrt{K}$ as well.

\subsection{Proof of Lemma~\ref{Gradient_statistical}}\label{proof:proof of lemma Gradient_statistical}

\begin{itemize}

	
	\item \textbf{The FCN case:}	
	Following from \eqref{eq:gradient_formula}, we have
	\begin{align*}
	\left\langle \nabla \ell \left(\bW \right) , \bu \right\rangle  = \frac{1}{K}\sum_{j=1}^{K} \left(\frac{\left( y-H\left(\bW \right) \right)\cdot \phi^{\prime} \left(\bw_{j}^{\top}\bx \right) }{H\left(\bW \right)\left(1-H\left(\bW \right) \right) }\right) \left(\bu_{j}^{\top}\bx \right), 
	\end{align*}	
	and by definition, we can upper-bound the sub-Gaussian norm as
	\begin{align*}
	\|\left\langle \nabla \ell \left(\bW \right), \bu \right\rangle\|_{\psi_{2}}  &\leq 
	\begin{cases}
	\frac{1}{K}\sum_{j=1}^{K} \left\|\frac{\phi^{\prime} \left(\bw_{j}^{\top}\bx \right)}{\left(1-\frac{1}{K}\sum_{l=1}^{K}\phi\left(\bw_{l}^{\top}\bx \right) \right)}\bu_{j}^{\top}\bx\right\|_{\psi_{2}}  \leq \sum_{j=1}^{K} \|\bu_{j}^{\top}\bx\|_{\psi_{2}}  & y = 0\\
	\frac{1}{K}\sum_{j=1}^{K} \left\|\frac{\phi^{\prime} \left(\bw_{j}^{\top}\bx \right)}{\frac{1}{K}\sum_{l=1}^{K}\phi\left(\bw_{l}^{\top}\bx \right)}\bu_{j}^{\top}\bx \right\|_{\psi_{2}}  \leq \sum_{j=1}^{K} \|\bu_{j}^{\top}\bx\|_{\psi_{2}}  & y =1
	\end{cases}.
	\end{align*}
	Thus we conclude that
	\begin{align}
	\|\left\langle \nabla \ell \left(\bW \right)  , \bu \right\rangle\|_{\psi_{2}} \leq  \sum_{j=1}^{K} \|\bu_{j}\|_{2} \leq   \sqrt{K},
	\end{align}
	and the directional gradient is $\sqrt{K}$-sub-Gaussian.

	
	\item \textbf{The CNN case:}
	Following from \eqref{eq:gradient_formula_CNN}, we have
	\begin{align*}
	\left\langle \nabla \ell\left(\bw \right),\bu  \right\rangle = -\sum_{j=1}^{K} \frac{1}{K} \phi^{\prime}\left(\bw^{\top}\bx^{\left(j \right) } \right) \frac{y-H\left( \bw\right) }{H\left( \bw\right)\left(1-H\left( \bw\right) \right) } \cdot \left(\bu^{\top}\bx^{\left(j \right) } \right),  
	\end{align*} 
	where 
	\begin{align*}
	\left|\phi^{\prime}\left(\bw^{\top}\bx^{\left(j \right) }\right) \frac{y-H\left( \bw\right)}{H\left( \bw\right)\left(1-H\left( \bw\right) \right) } \right| = \begin{cases}
	\frac{\phi^{\prime}\left(\bw^{\top}\bx^{\left(j \right) } \right)}{\sum_{j=1}^{K}\frac{1}{K}\phi\left(\bw^{\top}\bx^{\left(j \right) } \right) }\leq K & y=1 \\
	\frac{\phi^{\prime}\left(\bw^{\top}\bx^{\left(j \right) } \right)}{\sum_{j=1}^{K}\frac{1}{K}\left( 1-\phi\left(\bw^{\top}\bx^{\left(j \right) }\right)\right)  } \leq K & y = 0
	\end{cases}.  
	\end{align*}
	Then the sub-Gaussian norm of $\left\langle \nabla \ell\left(\bw \right),\bu  \right\rangle$ is upper bounded as
	\begin{align}
	\|\left\langle \nabla \ell\left(\bw \right),\bu  \right\rangle\|_{\psi_{2}} \leq K\cdot \frac{1}{K}\sum_{j=1}^{K} \|\bu^{\top}\bx^{\left(j \right) }\|_{\psi_{2}} \leq C\cdot K.
	\end{align}
	Hence, the directional gradient is $K$-sub-Gaussian.
\end{itemize}


\section{Proof of Theorem \ref{theorem:initial_guarantee}}\label{Appendix:initial}

The proof contains two parts. Part (a) proves that the estimation of the direction of $\bW^{\star}$ is sufficiently accurate, which follows from the arguments similar to those in \cite{zhong17a} and is only briefly summarized below. Part (b) is different, where we do not require the homogeneous condition for the activation function, and instead, our proof is based on a mild condition in Assumption \ref{assumption:inverse}. We detail our proof in part (b).



(a) In order to estimate the direction of each $\bw_i$ for $i=1,\ldots,K$, \cite{zhong17a} showed that for the regression problem,  if the sample size $n \geq d \mathrm{poly}\left( K,\kappa,\zeta,\log d \right) $, where $\zeta>1$ is any constant, then 
	\begin{equation}
		\|\overline{ \bw_i}^{\star}- s_i \bV \widehat{\bu}_i \| \leq \epsilon \mathrm{poly\left(K,\kappa \right) }
	\end{equation}
holds with probability at least $1-d^{-\Omega\left(\zeta \right) }$. Such a result also holds for the classification problem with only slight difference in the proof as we describe as follows. The main idea of the proof is to bound the estimation error of $\bP_{2}$ and $\bR_{3}$ via Bernstein inequality. For the regression problem, Bernstein inequality was applied to terms associated with each neuron individually, and the bounds were then put together via the triangle inequality in \cite{zhong17a}. However, for the classification problem here, we apply Bernstein inequality to the terms associated with all neurons together. Another difference is that the label $y_i$ of the classification model is bounded by nature, whereas the output $y_i$ in the regression model needs to be upper-bounded via homogeneously bounded conditions of the activation function. A reader can refer to \cite{zhong17a} for the details of the proof for this part.

(b) In order to estimate $\|\bw_{i}\|$ for $i=1,\ldots,K$, we provide a different proof from \cite{zhong17a}, which does not require the homogeneous condition on the activation function, but assumes a more relaxed condition in Assumption \ref{assumption:inverse}.

%

We define a quantity $Q_{1}$ as follows:
\begin{equation}\label{eq:def_Q1}
Q_{1} = \bM_{l_{1}} (\bI,\underbrace{\balpha,\cdots,\balpha}_{\left( l_{1}-1\right) } ),  
\end{equation} 
where $l_{1}$ is the first non-zero index such that $\bM_{l_{1}}\neq 0$. For example, if $l_{1}=3$, then $Q_{1}$ takes the following form
\begin{equation}
Q_{1} = \bM_{3}\left(\bI,\balpha,\balpha \right) = \frac{1}{K}\sum_{i=1}^{K} m_{3,i}(\|\bw_{i}^{\star}\|) \left(\balpha^{\top}\overline{ \bw}_{i}^{\star} \right)^{2} \overline{\bw}_{i}^{\star} ,
\end{equation}
where $\overline{ \bw} = \bw/\| \bw\|$ and by definition
\begin{equation}
m_{3,i} (\|\bw_{i}^{\star}\|)= \mathbb{E}\left[ \phi\left(\|\bw_{i}^{\star}\|\cdot z \right) z^{3}  \right] - 3 \mathbb{E}\left[ \phi\left(\|\bw_{i}^{\star}\|\cdot z \right) z\right].
\end{equation}


Clearly, $Q_1$ has information of $\|\bw_{i}^{\star}\|$, which can be estimated by solving the following optimization problem: 
\begin{equation}\label{z_star}
\beta^{\star} = \mathrm{argmin}_{\beta\in \mathbb{R}^{K}} \left\|\frac{1}{K}\sum_{i=1}^{K} \beta_{i} s_{i} \overline{ \bw_i}^{\star} - Q_{1}\right\|,
\end{equation}
where each entry of the solution takes the form 
\begin{align}\label{eq:beta_w}
\beta_{i}^{\star} = s_{i}^{3} m_{3,i}(\|\bw_{i}^{\star}\|) \left( \balpha^{T} s_{i} \overline{ \bw_i}^{\star} \right)^{2}.
\end{align} 
In the initialization, we substitute $\widehat{Q}_{1}$ (estimated from training data) for $Q_{1}$, $\bV \hat{u}_{i}$ (estimated in part (a)) for $s_{i}\overline{ \bw_i}^{\star} $ into \eqref{z_star}, and obtain an estimate $\hat{\beta}$ of $\beta^{\star}$. We then substitute $\hat{\beta}$ for $\beta^{\star}$ and $\bV \hat{u}_{i}$ for $s_{i}\overline{ \bw_i}^{\star} $ into \eqref{eq:beta_w}  to obtain an estimate $\hat{a}_i$ of $\|\bw_{i}^{\star}\|$ via the following equation
\begin{equation}\label{eq:mag_w}
\hat{\beta}_{i} = s_{i}^{3} m_{3,i} (\hat{a}_i) \left( \balpha^{\top} \bV \hat{u}_{i} \right)^{2}.	
\end{equation} 
Furthermore, since $m_{l_{1},i}(x)$ has fixed sign for $x>0$ and for $l_{1}\geq 1$, $s_{i}$ can be estimated correctly from the sign of $\hat{\beta}_{i}$ for $i=1,\ldots,K$.

For notational simplicity, let $\beta_{1,i}^{\star}:=\frac{\beta_{i}^{\star}}{s_{i}^{3}\left( \balpha^{\top} s_{i} \overline{ \bw_i}^{\star} \right)^{2}}$ and $\hat{\beta}_{1,i}:=\frac{\hat{\beta}_{i}}{s_{i}^{3}\left( \balpha^{\top} \bV \hat{u}_{i} \right)^{2}}$, and then \eqref{eq:beta_w} and \eqref{eq:mag_w} become
\begin{align}\label{eq:beta1}
\hat{\beta}_{1,i}=m_{3,i} (\hat{a}_i), \quad \beta_{1,i}^{\star} = m_{3,i}(\|\bw_{i}^{\star}\|) . 
\end{align}

By Assumption \ref{assumption:inverse} and \eqref{eq:beta_w}, there exists a constant $\delta'>0$ such that the inverse function $g(\cdot)$ of $m_{3,1}(\cdot)$ has upper-bounded derivative in the interval $(\beta_{1,i}^{\star}-\delta',\beta_{1,i}^{\star}+\delta')$, i.e., $|g'(x)|<\Gamma$ for a constant $\Gamma$. By employing the result in \cite{zhong17a}, if the sample size $n \geq d \mathrm{poly}\left( K,\kappa,t,\log d \right) $, then $\widehat{Q}_{1}$ and $Q_{1}$, $\bV \hat{u}_{i}$ and $s_{i}\overline{ \bw_i}^{\star} $ can be arbitrarily close so that $|\beta_{1,i}^{\star}-\hat{\beta}_{1,i}|<\min\{\delta',\frac{r}{\sqrt{K}\Gamma}\}$. 

Thus, by \eqref{eq:beta1} and the mean value theorem, we obtain
\begin{equation}
|\hat{a}_{i} - \|\bw_{i}^{\star}\|| = |g^{\prime}(\xi)||\beta_{1,i}^{\star}-\hat{\beta}_{1,i}|, 
\end{equation}
where $\xi$ is between $\beta_{1,i}^{\star}$ and $\hat{\beta}_{1,i}$, and hence $|g'(\xi)|<\Gamma$. Therefore,
$|\hat{a}_{i} - \|\bw_{i}^{\star}\|| \leq \frac{r}{\sqrt{K}}$, which is the desired result.

\end{document}